%% file: FMRVR.tex
\documentclass{rQUF2e}

\usepackage[stable,perpage,hang]{footmisc} % footnotes in \section,  reset footnote numbering for each page
\usepackage{upgreek}
\usepackage[ruled,linesnumbered]{algorithm2e}
\SetAlFnt{\footnotesize} % font size of algorithm
\numberwithin{equation}{section} % grouped equation numbering
\usepackage{subcaption}
\usepackage{tabularx}
\usepackage{makecell} % multi-row of table
\usepackage[space]{grffile} % include graphics with spaces
\usepackage[colorlinks=true, allcolors=blue, pdfborder={0 0 0}]{hyperref}
\usepackage{microtype} % because of 'fi'
\DisableLigatures{encoding=*,family=*} % because of 'fi'

\newcommand{\rmnum}[1]{\romannumeral #1} % Roman number
\newcommand{\defeq}{\buildrel\text{def}\over=} % define

\begin{document}
\title{Fast multi-output relevance vector regression %\thanks{Grants or other notes
}


%\author{Youngmin Ha}
%
%\institute{Y. Ha \at
%              Adam Smith Business School \\
%              University of Glasgow, UK \\
%              \email{y.ha.1@research.gla.ac.uk}           %  \\
%%             \emph{Present address:} of F. Author  %  if needed

\author{Youngmin Ha$^{\ast}$\thanks{$^\ast$Email: \href{mailto:y.ha.1@research.gla.ac.uk}{y.ha.1@research.gla.ac.uk}}}
%y.ha.1@research.gla.ac.uk}}
\affil{Adam Smith Business School, University of Glasgow, UK}

\maketitle
%\begin{linenumbers}
\begin{abstract}
This paper aims to decrease the time complexity of multi-output relevance vector regression from $O\left(VM^3\right)$ to $O\left(V^3+M^3\right)$, where $V$ is the number of output dimensions, $M$ is the number of basis functions, and $V<M$. The experimental results demonstrate that the proposed method is more competitive than the existing method, with regard to computation time. MATLAB codes are available at \url{http://www.mathworks.com/matlabcentral/fileexchange/49131}.
%\keywords{Relevance vector regression \and Relevance vector machine \and Sparse Bayesian learning}
\end{abstract}


\section{Introduction}
When it comes to multi-input nonparametric nonlinear regression or classification, the following three methods can be considered: support vector machine (SVM), relevance vector machine (RVM), and Gaussian process (GP) regression or classification. %They are all supervised learning methods for classification and regression: i.e.~training from data is required.

SVM, invented by \cite{cortes1995support}, is a popular machine learning tool. It uses kernel trick (RVM and GP also use the kernel trick) and makes classification and regression computationally efficient for multidimensional inputs. However, its disadvantage is that a user needs to choose a proper value of the error/margin trade-off parameter `$C$' (the proper value can be found by k-fold cross-validation).

RVM, invented by \cite{tipping2001sparse},\footnote{Single-output relevance vector regression is easily explained in \url{http://static1.squarespace.com/static/58851af9ebbd1a30e98fb283/t/58902f4a6b8f5ba2ed9d3bfe/1485844299331/RVM+Explained.pdf}.} avoids estimating the error/margin trade-off parameter `$C$' of SVM (and in regression, the additional insensitivity parameter `$\varepsilon$'~\citep{vapnik2000nature}, or `$\nu$'~\citep{scholkopf2000new}), which wastes computation time. Moreover, RVM allows probabilistic predictions (i.e.~both mean and variance of a Gaussian distribution) although SVM predicts only mean values (the error bar estimation of SVM is possible with additional computation~\citep{gao2002probabilistic,chu2004bayesian}).

GP regression or classification, invented by~\cite{gibbs1997bayesian}, also does not need estimating `$C$' (and the additional parameter of regression `$\varepsilon$' or `$\nu$'). Furthermore, the predictive variance of GP regression or classification changes over an input vector $\mathbf x_*$: the predictive variance is smaller at the denser region of training samples, while the predictive variance of RVM is almost constant over $\mathbf x_*$.

%The prediction of the Gaussian process model output value is given as a normal distribution, expressed in terms of mean and variance. A GP model is a probabilistic, nonparametric model for the prediction of output variable distributions. A user chooses the covariance kernel prior on the noisy observations. Gaussian process regression aims to reconstruct the underlying signal f by removing the contaminating noise ε.

Support vector regression (SVR), relevance vector regression (RVR), and GP regression are for multi-input but single-output regression, and they have been extended as multi-input and multi-output regression to model correlated outputs: multi-output SVR~\citep{perez2002multi,vazquez2003multi,tuia2011multioutput}, multi-output RVR~\citep{thayananthan2005template,thayananthan2008pose}, and multi-output GP regression~\citep{boyle2004dependent,bonilla2007multi,alvarez2009sparse}.

The multi-output relevance vector regression (MRVR) algorithm by~\citet[Chapter~6]{thayananthan2005template}, \citet{thayananthan2008pose} uses the Bayes' theorem and the kernel trick to perform regression, but it has the limitation of low computational efficiency. Therefore, a new faster algorithm is proposed in this paper: it uses the matrix normal distribution to model correlated outputs, while the existing algorithm uses the multivariate normal distribution.

The contributions of this paper are:
\begin{itemize}
	\item in Section~\ref{sec:pro}, to propose a new algorithm with less time complexity than the existing MRVR algorithm by~\citet[Chapter~6]{thayananthan2005template}, \citet{thayananthan2008pose};
	\item in Section~\ref{sec:results}, to present Monte Carlo simulation results to compare between the existing and the proposed MRVR algorithm in terms of accuracy and computation time.
\end{itemize}

The rest of this paper is organised as follows: Section~\ref{sec:related} lists related work. Section~\ref{sec:exi} and Section~\ref{sec:pro} describe the existing and proposed algorithms of MRVR, respectively. Section~\ref{sec:results} shows the experimental results by using MATLAB, and Section~\ref{sec:conclusion} gives the conclusion.

\section{Related work}\label{sec:related}
The computation time of the single-output SVM has decreased: \cite{guo2007reducing} reduced the SVR training time by reducing the number of training examples. \cite{catanzaro2008fast} accelerated the SVM computation by using both faster sequential algorithms and parallel computation on a graphics processing unit (GPU). \cite{chang2011libsvm} made fast and efficient C++ software of SVM.

The computation time of the single-output RVM has decreased: \cite{tipping2003fast} proposed a new marginal likelihood maximisation algorithm with efficient addition/deletion of candidate basis functions. \cite{ben2006accelerating} partitioned training samples into small chunks to avoid the inverse of a large matrix (the matrix inversion is the most computationally expensive part of RVM). \cite{yang2010high} accelerated RVM computation by parallelising the matrix inversion operation on a GPU.

The computation time of the single-output GP regression has decreased. \cite{seeger2003fast} reduced GP regression training time by approximating the likelihood of training data. \cite{shen2006fast} reduced both the training and prediction time of GP regression by using an approximation method, based on a tree-type multiresolution data structure. \cite{srinivasan2010gpuml} accelerated linear algebra operations of GP regression on a GPU, and \cite{gramacy2014massively} made a GPU accelerated version of GP regression approximation.

\section{Existing method}\label{sec:exi}
The following subsections describe the existing method of MRVR~\cite[Chapter~6]{thayananthan2005template}, \citep{thayananthan2008pose}.

\subsection{Model specification}\label{sec:exi:model}
$V$-dimensional multi-output regression upon an input $\mathbf x\in\mathbb R^{U\times 1}$ (\mbox{$U$-dimensional} column vector), a weight $\mathbf W\in\mathbb R^{M\times V}$ ($M$-by-$V$ matrix), and an output function $\mathbf y(\mathbf x;\mathbf W)\in\mathbb R^{1\times V}$ ($V$-dimensional row vector) (upright bold lower case letters denote vectors, and upright bold capital letters denote matrices) is expressed as
\begin{equation}
\mathbf y(\mathbf x;\mathbf W)=\left(\mathbf W^\intercal{\bm\upphi(\mathbf x)}\right)^\intercal,
\end{equation}
where $\mathbf W^\intercal$ is the transpose of the matrix $\mathbf W$ (the objective of this paper is to estimate proper values of $\mathbf W$), and \mbox{$\bm\upphi(\mathbf x)=[\phi_1(\mathbf x)~\phi_2(\mathbf x)~\ldots~\phi_M(\mathbf x)]^\intercal\in\mathbb R^{M\times 1}$} is the transformed input by nonlinear and fixed basis functions. In other words, the output $\mathbf y(\mathbf x;\mathbf W)$ is a linearly weighted sum of the transformed input $\bm\upphi(\mathbf x)$.

Given a data set of input-target pairs $\left\{\mathbf x_i\in\mathbb R^{U\times 1},\mathbf t_i\in\mathbb R^{1\times V}\right\}_{i=1}^N$, where $N$ is the number of training samples, it is assumed that the targets $\mathbf t_i$ are samples from the model $\mathbf y(\mathbf x_i;\mathbf W)$ with additive noise:
\begin{equation}\label{eq:exi:t}
\mathbf t_i=\mathbf y(\mathbf x_i;\mathbf W)+\bm\upepsilon_i,
\end{equation}
where $\mathbf W\in\mathbb R^{(N+1)\times V}$ is the weight,  $\bm\upepsilon_i\in\mathbb R^{1\times V}$ are independent samples from a Gaussian noise process with mean zero and a covariance matrix $\mathbf\Omega\in\mathbb R^{V\times V}$, and $\mathbf\Omega$ is decomposed as the diagonal matrix of the variances $\mathbf D\in\mathbb R^{V\times V}$ and the correlation matrix $\mathbf R\in\mathbb R^{V\times V}$:
\begin{equation}
\mathbf\Omega=\mathbf D^\frac{1}{2}\mathbf R\mathbf D^\frac{1}{2},
\end{equation}
where $\mathbf D=\textnormal{diag}\left(\sigma_1^2,\sigma_2^2,\ldots,\sigma_V^2\right)$, and $\sigma_j^2$ is the variance of the $j$-th dimension's noise. %For simple math, $\beta_j$ is substituted for $\sigma_j^{-2}$ in the subsequent expressions.

Because of the ignorance of $\mathbf R$ by~\cite{thayananthan2005template} and the assumption of independent Gaussian noise, the likelihood of the data set can be given by the product of the Gaussian distributions:
\begin{equation}\label{eq:exi:p(T|W,s)}
p\left(\mathbf T|\mathbf W,\bm\upsigma\right)=\prod_{j=1}^{V}\left(2\pi \sigma_j^2\right)^{-\frac{N} 2}\textnormal{exp}\left(-\frac{1}{2\sigma_j^2}\left\|\bm\uptau_j-\mathbf\Phi\mathbf w_j\right\|^2\right),
\end{equation}
where $\mathbf T=\left[\begin{matrix}
	\mathbf t_1\\
	\mathbf t_2\\
	\vdots\\
	\mathbf t_N
\end{matrix}\right]\in\mathbb R^{N\times V}$, $\bm\upsigma=[\sigma_1~\sigma_2~\ldots~\sigma_V]\in\mathbb R_{\ge0}^{1\times V}$, $\bm\uptau_j\in\mathbb R^{N\times 1}$ is the $j$-th column vector of $\mathbf T$, $\mathbf w_j\in\mathbb R^{(N+1)\times 1}$ is the $j$-th column vector of $\mathbf W$, \mbox{$\mathbf\Phi =[\bm\upphi
(\mathbf x_1)~\bm\upphi(\mathbf x_2)~\ldots~\bm\upphi
(\mathbf x_N)]^\intercal\in\mathbb R^{N\times (N+1)}$} is a design matrix, $\bm\upphi
(\mathbf x)=[1~K(\mathbf x,\mathbf x_1)~\ldots~K(\mathbf x,\mathbf x_N)]^\intercal\in\mathbb R^{(N+1)\times 1}$, and $K(\mathbf x,\mathbf x')$ is a kernel function. For
clarity, the implicit conditioning on the input $\mathbf x_i,\forall i$ is omitted in Eq.~(\ref{eq:exi:p(T|W,s)}) and the subsequent expressions.

An assumption to avoid over-fitting in estimation of $\mathbf W$ is
\begin{equation}
p\left(\mathbf W|\bm\upalpha\right)=\prod_{j=1}^{V}\prod_{i=0}^{N}\mathcal{N}\left(0,\alpha_i^{-1}\right).
\end{equation}
This means the prior distribution of $\mathbf w_j$ is zero-mean Gaussian with inverse variances \mbox{$\bm\upalpha=\left[\alpha_0~\alpha_1~\ldots~\alpha_N\right]^\intercal\in\mathbb R_{>0}^{(N+1)\times 1}$}, which are $N+1$ hyperparameters~\citep{tipping2001sparse}, and $\mathbf w_j$ and $\mathbf w_{j'}$ $(j\neq j')$ have the same distribution as $\prod_{i=0}^{N}\mathcal{N}\left(0,\alpha_i^{-1}\right)$.

\subsection{Inference}
By both the Bayes' theorem and the property of $p\left(\mathbf T|\mathbf W,\bm\upalpha,\bm\upsigma\right)=p\left(\mathbf T|\mathbf W,\bm\upsigma\right)$,\footnote{In the case that the weight $\mathbf W$ is given, its inverse variances $\bm\upalpha$ are redundant in the calculation of the conditional probability of the target $\mathbf T$.} the posterior probability distribution function over $\mathbf W$ is decomposed as
\begin{equation}
p\left(\mathbf W|\mathbf T,\bm\upalpha,\bm\upsigma \right)=\frac{p\left(\mathbf T|\mathbf W,\bm\upsigma \right)p\left(\mathbf W|\bm\upalpha,\bm\upsigma\right)}{p\left(\mathbf T|\bm\upalpha,\bm\upsigma\right)},
\end{equation}
and it is given by the product of multivariate Gaussian distributions:
\begin{equation}\label{eq:exi:p(W|T,a,s)}
p\left(\mathbf W|\mathbf T,\bm\upalpha,\bm\upsigma \right)=\prod_{j=1}^{V}\left(2\pi\right)^{-\frac{N+1} 2}|\mathbf\Sigma_j|^{-\frac{1}{2}}\textnormal{exp}\left(-\frac{1}{2}\left(\mathbf w_j-\bm\upmu_j\right)^\intercal\mathbf\Sigma_j^{-1}\left(\mathbf w_j-\bm\upmu_j\right)\right),
\end{equation}
where the $j$-th posterior covariance and mean are, respectively:
\begin{equation}\label{eq:exi:Sigma}
\mathbf\Sigma_j =\left(\sigma_j^{-2}\mathbf\Phi^\intercal\mathbf\Phi +\mathbf A\right)^{-1},
\end{equation}
\begin{equation}\label{eq:exi:mu}
\bm\upmu_j=\sigma_j^{-2}\mathbf\Sigma_j \mathbf\Phi^\intercal\bm\uptau_j,
\end{equation}
where $\mathbf A=\textnormal{diag}\left(\alpha_0,\alpha_1,\ldots,\alpha_N\right)\in\mathbb R^{(N+1)\times(N+1)}$.

In the case of uniform hyperpriors $\bm\upalpha$ and $\bm\upsigma$, maximising a posteriori \mbox{$p\left(\bm\upalpha,\bm\upsigma\right|\mathbf T)\propto p\left(\mathbf T|\bm\upalpha,\bm\upsigma\right)p\left(\bm\upalpha\right)p\left(\bm\upsigma\right)$} is equivalent to maximising the marginal likelihood $p\left(\mathbf T|\bm\upalpha,\bm\upsigma\right)$, which is given by
\begin{equation}\label{eq:exi:p(T|a,s)}
p\left(\mathbf T|\bm\upalpha,\bm\upsigma\right)=\prod_{j=1}^{V}\left(2\pi
\right)^{-\frac{N} 2}\left|\sigma_j^2\mathbf I+\mathbf\Phi\mathbf A^{-1}\mathbf\Phi ^\intercal\right|^{-\frac{1}
2}\textnormal{exp}\left(-\frac{1} 2\bm\uptau_j^\intercal\left(\sigma_j^2\mathbf I+\mathbf\Phi \mathbf A^{-1}\mathbf\Phi^\intercal\right)^{-1}\bm\uptau_j\right).
\end{equation}

\subsection{Marginal likelihood maximisation}
The same method of accelerating the univariate relevance vector machine~\citep{tipping2003fast} is used to accelerate the existing algorithm.

The log of Eq.~(\ref{eq:exi:p(T|a,s)}) is an objective function:
\begin{equation}\label{eq:exi:L(a,s)}
\mathcal L(\bm\upalpha,\bm\upsigma)=-\frac{1} 2\sum_{j=1}^{V}\left(N\log(2\pi)+\log
\left|\mathbf C_j\right|+\bm\uptau_j^\intercal \mathbf C_j^{-1}\bm\uptau_j\right),
\end{equation}
where $\mathbf C_j=\sigma_j^2\mathbf I+\mathbf\Phi \mathbf A^{-1}\mathbf\Phi ^\intercal\in\mathbb R^{N\times N}$, and by considering the dependence of $\mathcal L(\bm\upalpha,\bm\upsigma)$ on a single hyperparameter $\alpha_i,i\in\{0,1,\ldots,N\}$, $\mathbf C_j$ is decomposed as the following two parts:
\begin{equation}\begin{split}
\mathbf C_j&=\sigma_j^2\mathbf I+\underset{m\neq i}{\sum }\alpha_m^{-1}\bm\upphi_m\bm\upphi_m^\intercal+\alpha_i^{-1}\bm\upphi_i\bm\upphi_i^\intercal\\
&=\mathbf C_{-i,j}+\alpha_i^{-1}\bm\upphi_i\bm\upphi_i^\intercal,
\end{split}\end{equation}
where $\mathbf C_{-i,j}\in\mathbb R^{N\times N}$ is $\mathbf C_j$ with the contribution of a basis vector $\bm\upphi _i\in\mathbb R^{N\times 1}$ removed, and
\begin{equation}
\bm\upphi_i=\left\{\begin{array}{cl}
[1~1~\ldots~1]^\intercal,&\mbox{ if $i=0$} \\
~[K(\mathbf x_i,\mathbf x_1)~K(\mathbf x_i,\mathbf x_2)~\ldots~K(\mathbf x_i,\mathbf x_N)]^\intercal,&\mbox{ otherwise}
\end{array}\right..
\end{equation}
The determinant and inverse matrix of $\mathbf C_j$ are, respectively:
\begin{equation}
\left|\mathbf C_j\right|=\left|\mathbf C_{-i,j}\right|\left(1+\alpha _i^{-1}\bm\upphi_i^\intercal\mathbf C_{-i,j}^{-1}\bm\upphi_i\right),
\end{equation}
by Sylvester's determinant theorem, and
\begin{equation}
\mathbf C_j^{-1}=\mathbf C_{-i,j}^{-1}-\frac{\mathbf C_{-i,j}^{-1}\bm\upphi_i\bm\upphi_i^\intercal\mathbf C_{-i,j}^{-1}}{\alpha _i+\bm\upphi_i^\intercal\mathbf C_{-i,j}^{-1}\bm\upphi_i},
\end{equation}
by Woodbury matrix identity. From these, $\mathcal L(\bm\upalpha,\bm\upsigma)$ in Eq.~(\ref{eq:exi:L(a,s)}) can be decomposed into $\mathcal L(\bm\upalpha_{-i},\bm\upsigma)$, the marginal likelihood with $\bm\upphi_i$ excluded, and $\ell(\alpha_i,\bm\upsigma)$, the isolated marginal likelihood of $\bm\upphi_i$:
\begin{equation}\begin{split}
\mathcal L(\bm\upalpha,\bm\upsigma)=&-\frac{1} 2\sum_{j=1}^{V}\left(N\log(2\pi) +\log\left|\mathbf C_{-i,j}\right|+\bm\uptau_j^\intercal \mathbf C_{-i,j}^{-1}\bm\uptau_j\right)\\
&-\frac 1 2\sum_{j=1}^{V}\left(-\log\alpha _i+\log\left(\alpha_i+\bm\upphi_i^\intercal\mathbf C_{-i,j}^{-1}\bm\upphi_i\right)-\frac{\left(\bm\upphi_i^\intercal\mathbf C_{-i,j}^{-1}\bm\uptau_j\right)^2}{\alpha _i+\bm\upphi_i^\intercal\mathbf C_{-i,j}^{-1}\bm\upphi_i}\right)\\
=&\mathcal L(\bm\upalpha_{-i},\bm\upsigma)+\frac 1 2\sum_{j=1}^{V}\left(\log \alpha_i-\log \left(\alpha
_i+s_{i,j}\right)+\frac{q_{i,j}^2}{\alpha_i+s_{i,j}}\right)\\
=&\mathcal L(\bm\upalpha_{-i},\bm\upsigma)+\ell(\alpha_i,\bm\upsigma),
\end{split}\end{equation}
where $s_{i,j}$ and $q_{i,j}$ are defined as, respectively:
\begin{subequations}\label{eq:exi:sqOriginal}
	\begin{equation}
	s_{i,j}\defeq\bm\upphi_i^\intercal\mathbf C_{-i,j}^{-1}\bm\upphi_i,
	\end{equation}
	\begin{equation}
	q_{i,j}\defeq\bm\upphi_i^\intercal\mathbf C_{-i,j}^{-1}\bm\uptau_j.
	\end{equation}
\end{subequations}

To avoid the matrix inversion of $\mathbf C_{-i}$ in Eq.~(\ref{eq:exi:sqOriginal}), which requires the time complexity of $O\left(N^3\right)$, $s_{i,j}'$ and $q_{i,j}'$ are computed as, respectively (by the Woodbury matrix identity):\footnote{$s_{i,j}'=\sigma_j^{-2}\bm\upphi_i^\intercal\bm\upphi_i$ and $q_{i,j}'=\sigma_j^{-2}\bm\upphi_i^\intercal\bm\uptau_j$ when $\alpha_i=\infty,\forall i$.}
\begin{subequations}\label{eq:exi:sqPrime}
	\begin{equation}\begin{split}
	s_{i,j}'&=\bm\upphi_i^\intercal\mathbf C_j^{-1}\bm\upphi_i\\
	&=\sigma_j^{-2}\bm\upphi_i^\intercal\bm\upphi_i-\sigma_j^{-4}\bm\upphi_i^\intercal\mathbf\Phi \mathbf\Sigma_j\mathbf\Phi ^\intercal\bm\upphi_i,
	\end{split}\end{equation}
	\begin{equation}\begin{split}
	q_{i,j}'&=\bm\upphi_i^\intercal\mathbf C_j^{-1}\bm\uptau_j\\
	&=\sigma_j^{-2}\bm\upphi_i^\intercal\bm\uptau_j-\sigma_j^{-4}\bm\upphi_i^\intercal\mathbf\Phi \mathbf\Sigma_j\mathbf\Phi^\intercal\bm\uptau_j,
	\end{split}\end{equation}
\end{subequations}
and then $s_{i,j}$ and $q_{i,j}$ in Eq.~(\ref{eq:exi:sqOriginal}) are computed as, respectively:\footnote{$s_{i,j}=s_{i,j}'$ and $q_{i,j}=q_{i,j}'$ when $\alpha_i=\infty$.}
\begin{subequations}\label{eq:exi:sq}
	\begin{equation}
		s_{i,j}=\frac{\alpha_is_{i,j}'}{\alpha_i-s_{i,j}'},
	\end{equation}
	\begin{equation}
		q_{i,j}=\frac{\alpha_i q_{i,j}'}{\alpha_i-s_{i,j}'}.
	\end{equation}
\end{subequations}

$\mathcal L(\bm\upalpha,\bm\upsigma)$ has a unique maximum with respect to $\alpha_i$ when the following equation is satisfied:
\begin{equation}\label{eq:exi:round_l(a,s)}
\frac{\partial\ell(\alpha_i,\bm\upsigma)}{\partial\alpha_i}=\frac 1 2\sum_{j=1}^{V}\left(\frac{1}{\alpha_i}-\frac{1}{\alpha
_i+s_{i,j}}-\frac{q_{i,j}^2}{(\alpha_i+s_{i,j})^2}\right)=0,
\end{equation}
which is a $(2V-1)$-th order polynomial equation of $\alpha_i$. This implies that:
\begin{itemize}
  \item If $\bm\upphi_i$ is ``in the model" (i.e. $\alpha_i<\infty$) and $\alpha_i$ in Eq.~(\ref{eq:exi:round_l(a,s)}) has at least one positive real root ($\alpha_i>0$ as $\alpha_i$ is inverse variance); then, $\alpha_i$ is re-estimated,
  \item If $\bm\upphi_i$ is ``in the model" (i.e. $\alpha_i<\infty$) yet $\alpha_i$ in Eq.~(\ref{eq:exi:round_l(a,s)}) does not have any positive real root; then, $\bm\upphi_i$ may be deleted (i.e. $\alpha_i$ is set to be $\infty$),
  \item If $\bm\upphi_i$ is ``out of the model" (i.e. $\alpha_i=\infty$) yet $\alpha_i$ in Eq.~(\ref{eq:exi:round_l(a,s)}) has at least one positive real root; then, $\bm\upphi_i$ may be added (i.e. $\alpha_i$ is set to be a finite value).
\end{itemize}

In addition, $\displaystyle\frac{\partial\mathcal L(\bm\upalpha,\bm\upsigma)}{\partial\sigma_j^2}=0$ leads to that $\mathcal L(\bm\upalpha,\bm\upsigma)$ has a unique maximum with respect to $\sigma_j^2$ when:
\begin{equation}\label{eq:exi:sigma_j^2}
\sigma_j^2=\frac{\left\|\bm\uptau_j-\mathbf\Phi\bm\upmu_j\right\|^2}{N-\sum_{i=1}^{N+1}\gamma_{i,j}},
\end{equation}
where $\gamma_{i,j}\defeq1-\alpha_{(i-1)}\Sigma_{j,ii}$, and $\Sigma_{j,ii}$ is the $i$-th diagonal element of \mbox{$\mathbf\Sigma_j\in\mathbb R^{(N+1)\times(N+1)}$}.

\subsection{Expectation-maximisation (EM) algorithm}
%\begin{spacing}{-0.5}
\begin{algorithm}
\caption{Existing EM algorithm of MRVR.}\label{alg:exi}
\KwIn{$\mathbf T\in\mathbb R^{N\times V}$, and $\bm\upphi_i\in\mathbb R^{N\times 1},\forall i=\{0,1,\ldots,N\}$, where $N$ is the number of training samples, and $V$ is the number of output dimensions}
\KwOut{$\mathbf\Sigma_j\in\mathbb R^{M\times M},\bm\upmu_j\in\mathbb R^{M\times 1},$ and $\sigma_j,\forall j=\{1,2,\ldots,V\}$, where $M$ is the number of basis functions in the model}
\tcp{Initialisation}
$\alpha_i\gets\infty,\forall i=\{0,1,\ldots,N\}$\\
\For{$j\gets1~\text{to}~V$}{
	$\bar{t}_j\gets\frac{1}{N}\sum_{i=1}^{N}t_{i,j}$\\
	$\sigma_j^2\gets\frac{0.1}{N-1}\sum_{i=1}^{N}\left(t_{i,j}-\bar{t}_j\right)^2$\\
}
convergence$\gets$false, $n\gets1$, $M\gets0$, where $n$ is the iteration number, and $M$ is the number of basis functions.\\
\While{convergence=false}{
	\tcp{maximisation step}
	\For{$i\gets0~\text{to}~N$}{
		\For{$j\gets1~\text{to}~V$}{
			Update $s_{i,j}'$ and $q_{i,j}'$ using Eq.~(\ref{eq:exi:sqPrime}), and 
			Update $s_{i,j}$ and $q_{i,j}$ using Eq.~(\ref{eq:exi:sq}).\\
		}
		\Switch{the number of positive real roots of Eq.~(\ref{eq:exi:round_l(a,s)})}{
			\uCase{0}{
				$\tilde\alpha_i\gets\infty$
			}
			\uCase{1}{
				$\tilde\alpha_i\gets$ the positive real root of Eq.~(\ref{eq:exi:round_l(a,s)})
			}
			\Other{
				$\tilde\alpha_i\gets$ one of the positive real roots of Eq.~(\ref{eq:exi:round_l(a,s)}), which maximises $2\Delta\mathcal L_i$ of either \rmnum1) Eq.~(\ref{eq:exi:2DLest}) if $\alpha_i<\infty$ or \rmnum2) Eq.~(\ref{eq:exi:2DLadd}) if $\alpha_i=\infty$ %\footnotemark[\ref{fnt:estOrAdd}]
			}
		}
		\uIf{$\tilde\alpha_i<\infty$}{
			\eIf($z_i\gets$`re-estimation'){$\alpha_i<\infty$}{
				Update $2\Delta\mathcal L_i$ using Eq.~(\ref{eq:exi:2DLest}).}
			($z_i\gets$`addition'){
				Update $2\Delta\mathcal L_i$ using Eq.~(\ref{eq:exi:2DLadd}).}
		}
		\uElseIf($z_i\gets$`deletion'){$\alpha_i<\infty$}{
			Update $2\Delta\mathcal L_i$ using Eq.~(\ref{eq:exi:2DLdel}).
		}
		\Else{$2\Delta\mathcal L_i\gets-\infty$}
	}
	$i\gets\arg\max_i2\Delta\mathcal L_i$~\tcp{Select $i$ which gives the greatest increase of the marginal likelihood}
	\If{$n\neq1$}{
		Update $\sigma_j,\forall j$ using Eq.~(\ref{eq:exi:sigma_j^2(new)}).
	}	
	\Switch{$z_i$}{
		\uCase{`re-estimation'}{
			$\Delta\log\alpha\gets\log\frac{\alpha_i}{\tilde\alpha_i}$\\
			$\alpha_i\gets\tilde\alpha_i$\\
			\tcp{Check convergence (convergence criteria are the same as those in~\citep{tipping2003fast})}
			\If{$|\Delta\log\alpha|<0.1$}
				{convergence$\gets$true\\
				\For{$i\gets0~\text{to}~N$}{
					\If(\tcp*[h]{if $\bm{\upphi}_i$ is "out of the model"}){$\alpha_i=\infty$}{
						\If(\tcp*[h]{if $\bm{\upphi}_i$ may be added}){$\tilde\alpha_i<\infty$}{convergence$\gets$false\\break for loop}
					}
				}
			}
	    }
	    \uCase{`addition'}{
	    	$\alpha_i\gets\tilde\alpha_i$, 
	    	$M\gets M+1$
	    }
	    \uCase{`deletion'}{
	    	$\alpha_i\gets\infty$, 
	    	$M\gets M-1$
	    }
	}
	
	\tcp{Expectation step}
	Sequentially update \rmnum1) $\mathbf\Phi\in\mathbb R^{N\times M}$, $\mathbf A\in\mathbb R^{M\times M}$, \rmnum2) $\mathbf\Sigma_j\in\mathbb R^{M\times M},\forall j$, and \rmnum3)~$\bm\upmu_j\in\mathbb R^M,\forall j$ using Eq.~(\ref{eq:exi:Sigma}) and Eq.~(\ref{eq:exi:mu}), where $\mathbf\Phi,~\mathbf\Sigma_j$, and $\bm\upmu_j$ contain only $M$ basis functions that are currently included in the model, and the diagonal matrix $\mathbf A$ consists of $M$ hyperparameters of $\alpha_i$ that are currently included in the model.\\
	$n\gets n+1$\\
}
\end{algorithm}
%\end{spacing}
%\stepcounter{footnote}\footnotetext{Eq.~(\ref{eq:exi:2DLest}) is used if $\alpha_i<\infty$, and Eq.~(\ref{eq:exi:2DLadd}) is used if $\alpha_i=\infty$.\label{fnt:estOrAdd}}

Algorithm~\ref{alg:exi}, an EM algorithm to maximise the marginal likelihood, starts without any basis vector (i.e. $M=0$) and selects the basis vector $\bm\upphi_i$ which gives the maximum change of the marginal likelihood $\mathcal L(\bm\upalpha,\bm\upsigma)$ of Eq.~(\ref{eq:exi:L(a,s)}) at every iteration.

For efficient computation of the EM algorithm, quantities $\mathbf\Phi\in\mathbb R^{N\times M}$, \mbox{$\mathbf\Sigma_j\in\mathbb R^{M\times M}$}, and \mbox{$\bm\upmu_j\in\mathbb R^{M\times 1}$} contain only $M$ ($M\le N+1$ is always satisfied) basis functions that are currently included in the model (i.e. $\bm\upphi_i$ satisfying $\alpha_i<\infty$), and the diagonal matrix $\mathbf A$ consists of $M$ hyperparameters of $\alpha_i$ that are currently included in the model (i.e. $\alpha_i$ satisfying $\alpha_i<\infty$). Additionally, Eq.~(\ref{eq:exi:sigma_j^2}) is rewritten as
\begin{equation}\label{eq:exi:sigma_j^2(new)}
\sigma_j^2=\frac{\left\|\bm\uptau_j-\mathbf\Phi\bm\upmu_j\right\|^2}{N-\sum_{i=1}^{M}\gamma_{i,j}'},
\end{equation}
where $\gamma_{i,j}'\defeq1-\alpha_i'\Sigma_{j,ii}$, $\alpha_i'$ is the $i$-th non-infinity value of $\bm\upalpha$, and $\Sigma_{j,ii}$ is the $i$-th diagonal element of $\mathbf\Sigma_j\in\mathbb R^{M\times M}$.

From Eq.~(\ref{eq:exi:L(a,s)}), the change in the marginal likelihood can be written as
\begin{equation}\begin{split}\label{eq:exi:2DL}
2\Delta\mathcal L&=2\left(\mathcal L(\tilde{\bm\upalpha},\bm\upsigma)-\mathcal L(\bm\upalpha,\bm\upsigma)\right)\\
&=\sum_{j=1}^{V}\left(\log\frac{\left|\mathbf C_j\right|}{\left|\tilde{\mathbf C_j}\right|}+\bm\uptau_j^\intercal\left(\mathbf C_j^{-1}-\tilde{\mathbf C}_j^{-1}\right)\bm\uptau_j\right),
\end{split}\end{equation}
where updated quantities are denoted by a tilde (e.g., $\tilde{\bm{\upalpha}}$ and $\tilde{\mathbf C}_j$). Eq.~(\ref{eq:exi:2DL}) differs
according to whether $\alpha_i$ is re-estimated, added, or deleted:
\paragraph{Re-estimation} as $\mathbf C_j=\mathbf C_{-i,j}+\alpha _i^{-1}\bm\upphi_i\bm\upphi_i^\intercal$ and $\tilde{\mathbf C}=\mathbf C_{-i,j}+\tilde\alpha_i^{-1}\bm\upphi_i\bm\upphi_i^\intercal$,
\begin{equation}\label{eq:exi:2DLest}
2\Delta\mathcal L_i=\sum_{j=1}^{V}\left(\frac{q_{i,j}'^2}{s_{i,j}'+\left(\tilde\alpha_i^{-1}-\alpha_i^{-1}\right)^{-1}}-\log \left(1+s_{i,j}'\left(\tilde\alpha_i^{-1}-\alpha_i^{-1}\right)\right)\right),
\end{equation}
where $\tilde\alpha_i$ is re-estimated $\alpha_i$,
\paragraph{Addition} as $\mathbf C_j=\mathbf C_{-i,j}$ and $\tilde{\mathbf C}_j=\mathbf C_{-i,j}+\tilde\alpha_i^{-1}\bm\upphi_i\bm\upphi_i^\intercal$,
\begin{equation}\label{eq:exi:2DLadd}
2\Delta\mathcal L_i=\sum_{j=1}^{V}\left(\frac{q_{i,j}^2}{\tilde\alpha_i+s_{i,j}}+\log\frac{\tilde\alpha_i}{\tilde\alpha_i+s_{i,j}}\right),
\end{equation}
\paragraph{Deletion} as $\mathbf C_j=\mathbf C_{-i,j}+\alpha _i^{-1}\bm\upphi_i\bm\upphi_i^\intercal$ and $\tilde{\mathbf C}_j=\mathbf C_{-i,j}$,
\begin{equation}\label{eq:exi:2DLdel}
2\Delta\mathcal L_i=\sum_{j=1}^{V}\left(\frac{q_{i,j}'^2}{s_{i,j}'-\alpha_i}-\log\left(1-\frac{s_{i,j}'}{\alpha_i}\right)\right).
\end{equation}

\subsection{Making predictions}\label{sec:exi:predict}
We can predict both the mean of $j$-th output dimension $y_{*,j}$ and its variance $\sigma_{*,j}^2$ from a new input vector $\mathbf x_*$ based on both \rmnum1) Eq.~(\ref{eq:exi:t}), the model specification, and \rmnum2) Eq.~(\ref{eq:exi:p(W|T,a,s)}), the posterior distribution over the weights, conditioned on the most probable (MP) hyperparameters: \mbox{$\bm\upalpha_\textnormal{MP}\in\mathbb R_{>0}^{M\times 1}$} and $\bm\upsigma_\textnormal{MP}\in\mathbb R_{\ge0}^{1\times V}$, obtained from Algorithm~\ref{alg:exi}. Predictive distribution of $t_{*,j}$ is normally distributed as 
\begin{equation}
p(t_{*,j}|\mathbf T,\bm{\upalpha}_\textnormal{MP},\bm\upsigma_\textnormal{MP})=\mathcal N\left(t_{*,j}|y_{*,j},\sigma_{*,j}^2\right),
\end{equation}
with
\begin{equation}
y_{*,j}=\bm\upphi(\mathbf x_*)^\intercal\bm\upmu_j,
\end{equation}
and
\begin{equation}\label{eq:exi:sigma_*^2}
\sigma_{*,j}^2=\sigma_{\textnormal{MP},j}^2+\bm\upphi(\mathbf x_*)^\intercal\mathbf\Sigma_j\bm\upphi(\mathbf x_*),
\end{equation}
where $\bm\upphi(\mathbf x_*)\in\mathbb R^{M\times 1}$ comes from only $M$ basis functions that are included in the model after the EM algorithm, and subscript $j$ refers to the $j$-th output dimension. The predictive variance $\sigma_{*,j}^2$ comprises the sum of two variance components: the estimated noise on the training data $\sigma_{\textnormal{MP},j}^2$ and that due to the uncertainty in the prediction of the weights $\bm\upphi(\mathbf x_*)^\intercal\mathbf\Sigma_j\bm\upphi(\mathbf x_*)$.

\subsection{Algorithm complexity}
Matrix inversion of $\mathbf\Sigma_j\in\mathbb R^{M\times M}$ in Eq.~(\ref{eq:exi:Sigma}) for all $j\in\{1,2,\ldots,V\}$ determines \rmnum1)~the time complexity of the existing algorithm as $O\left(VM^3\right)$ and \rmnum2) the memory complexity as $O\left(VM^2\right)$, where $V$ is the number of output dimensions, and $M$ is the number of basis functions.\footnote{The matrix multiplication to calculate $s_{i,j}'$ and $q_{i,j}'$ in Eq.~\eqref{eq:exi:sqPrime} for all $i\in\{1,2,\ldots,N\}$, $j\in\{1,2,\ldots,V\}$ at the 11-th line of Algorithm~\ref{alg:exi} has the same time complexity because the matrix multiplication $\mathbf\Phi \mathbf\Sigma_j\mathbf\Phi ^\intercal$ is pre-calculated. In other words, the time complexity of the matrix multiplication is $O\left(VM^3\right)$, not $O\left(NVM^3\right)$, because $\mathbf\Phi \mathbf\Sigma_j\mathbf\Phi ^\intercal$ is independent of $i$.}

\section{Proposed method}\label{sec:pro}
\subsection{Model specification}\label{sec:pro:model}
Given a data set of input-target pairs $\left\{\mathbf x_i\in\mathbb R^{U\times 1},\mathbf t_i\in\mathbb R^{1\times V}\right\}_{i=1}^N$, where $N$ is the number of training samples, it is assumed that the targets $\mathbf t_i$ are samples from the model $\mathbf y(\mathbf x_i;\mathbf W)$ with additive noise:
\begin{equation}\label{eq:pro:t_i}
\mathbf t_i=\mathbf y(\mathbf x_i;\mathbf W)+\bm\upepsilon_i,
\end{equation}
where $\mathbf W\in\mathbb R^{(N+1)\times V}$ is the weight and $\bm\upepsilon_i\in\mathbb R^{1\times V}$ are independent samples from a Gaussian noise process with mean zero and a covariance matrix $\mathbf\Omega\in\mathbb R^{V\times V}$.

Eq.~(\ref{eq:pro:t_i}) can be rewritten, using matrix algebra, as
\begin{equation}\label{eq:pro:T}
\mathbf T=\mathbf\Phi \mathbf W+\mathbf E,
\end{equation}
where $\mathbf T=\left[\begin{matrix}
	\mathbf t_1\\
	\mathbf t_2\\
	\vdots\\
	\mathbf t_N
 \end{matrix}\right]\in\mathbb R^{N\times V}$ is the target, $\mathbf E=\left[\begin{matrix}
 	\bm\upepsilon_1\\
 	\bm\upepsilon_2\\
 	\vdots\\
 	\bm\upepsilon_N
\end{matrix}\right]\in\mathbb R^{N\times V}$ is the noise, \mbox{$\mathbf\Phi =[\bm\upphi
(\mathbf x_1)~\bm\upphi(\mathbf x_2)~\ldots~\bm\upphi
(\mathbf x_N)]^\intercal\in\mathbb R^{N\times (N+1)}$} is a design matrix, \mbox{$\bm\upphi
(\mathbf x)=[1~K(\mathbf x,\mathbf x_1)~K(\mathbf x,\mathbf x_2)~\ldots~K(\mathbf x,\mathbf x_N)]^\intercal\in\mathbb R^{(N+1)\times 1}$}, and $K(\mathbf x,\mathbf x')$ is a kernel function.

Because of the assumption of independent Gaussian noise, the likelihood of the data set can be given by the matrix Gaussian distribution:
\begin{equation}\label{eq:pro:p(T|W,O)}
p\left(\mathbf T|\mathbf W,\mathbf\Omega \right)=\left(2\pi \right)^{-\frac{VN} 2}\left|\mathbf\Omega
\right|^{-\frac{N} 2}\textnormal{exp}\left(-\frac{1}{2}\textnormal{tr}\left(\mathbf\Omega ^{-1}\left(\mathbf T-\mathbf\Phi \mathbf W\right)^\intercal\left(\mathbf T-\mathbf\Phi \mathbf W\right)\right)\right),
\end{equation}
where $\mathbf\Omega=\frac{\mathbb E\left[\mathbf E^\intercal \mathbf E\right]}{N}$, and $\textnormal{tr}$ denotes trace.\addtocounter{footnote}{1}\footnote{If $\mathbf\Omega=\textnormal{diag}\left(\sigma_1^2,\sigma_2^2,\ldots,\sigma_V^2\right)$, then Eq.~\eqref{eq:pro:p(T|W,O)} will be Eq.~\eqref{eq:exi:p(T|W,s)}.} As I assumed, $\mathbf I=\frac{\mathbb E\left[\mathbf E\mathbf E^\intercal\right]}{\textnormal{tr}(\mathbf\Omega)}$, which means the noise is independent among rows with the same variance, where $\mathbf I$ is an $N\times N$ identity matrix. For
clarity, the implicit conditioning on the input $\mathbf x_i,\forall i$ is omitted in Eq.~(\ref{eq:pro:p(T|W,O)}) and the subsequent expressions.

An assumption to avoid over-fitting in the estimation of $\mathbf W$ is
\begin{equation}\label{eq:pro:p(W|a,O)}
p\left(\mathbf W|\bm\upalpha,\mathbf\Omega\right)=\left(2\pi \right)^{-\frac{V\left(N+1\right)}
2}\left|\mathbf\Omega \right|^{-\frac{N+1} 2}\left|\mathbf A \right|^{\frac{V} 2}\textnormal{exp}\left(-\frac{1} 2\textnormal{tr}\left(\mathbf\Omega
^{-1}\mathbf W^\intercal \mathbf A\mathbf W\right)\right),
\end{equation}
where $\mathbf A^{-1}=\textnormal{diag}\left(\alpha_0^{-1},\alpha_1^{-1},\ldots,\alpha_N^{-1}\right)=\frac{\mathbb E\left[\mathbf W\mathbf W^\intercal\right]}{\textnormal{tr}(\mathbf\Omega)}$.
This means the prior distribution of $\mathbf W$ is zero-mean Gaussian with among-row inverse variances \mbox{$\bm\upalpha=\left[\alpha_0~\alpha_1~\ldots~\alpha_N\right]^\intercal\in\mathbb R_{>0}^{(N+1)\times 1}$}, which are $N+1$ hyperparameters~\citep{tipping2001sparse}. Eq.~(\ref{eq:pro:p(W|a,O)}) implies another assumption: $\mathbf\Omega=\frac{\mathbb E\left[\mathbf W^\intercal \mathbf W\right]}{\textnormal{tr}\left(\mathbf A^{-1}\right)}$. Actually, this is unreasonable because the weight $\mathbf W$ has no relationship with the noise $\mathbf E$ (i.e.~$\mathbf I=\frac{\mathbb E\left[\mathbf W^\intercal \mathbf W\right]}{\textnormal{tr}\left(\mathbf A^{-1}\right)}$, which means that the weights of different output dimensions are not correlated, is a reasonable assumption), but it is essential for creating a computationally efficient algorithm.

\subsection{Inference}
By both the Bayes' theorem and the property of $p\left(\mathbf T|\mathbf W,\bm\upalpha,\mathbf\Omega \right)=p\left(\mathbf T|\mathbf W,\mathbf\Omega \right)$,\footnote{In the case that the weight $\mathbf W$ is given, its inverse variances $\bm\upalpha$ are redundant in the calculation of the conditional probability of the target $\mathbf T$.} the posterior probability distribution function over $\mathbf W$ is decomposed as
\begin{equation}
p\left(\mathbf W|\mathbf T,\bm\upalpha,\mathbf\Omega \right)=\frac{p\left(\mathbf T|\mathbf W,\mathbf\Omega \right)p\left(\mathbf W|\bm\upalpha,\mathbf\Omega\right)}{p\left(\mathbf T|\bm\upalpha,\mathbf\Omega\right)},
\end{equation}
and it is given by the matrix Gaussian distribution:\addtocounter{footnote}{1}\footnote{\label{fnt:App}The proof is in Appendix.}
\begin{equation}\label{eq:pro:p(W|T,a,O)}
p\left(\mathbf W|\mathbf T,\bm\upalpha,\mathbf\Omega\right)=\left(2\pi \right)^{-\frac{V\left(N+1\right)} 2}\left|\mathbf\Omega \right|^{-\frac{N+1} 2}\left|\mathbf\Sigma \right|^{-\frac{V}
2}\textnormal{exp}\left(-\frac{1} 2\textnormal{tr}\left(\mathbf\Omega ^{-1}\left(\mathbf W-\mathbf M\right)^\intercal\mathbf\Sigma ^{-1}\left(\mathbf W-\mathbf M\right)\right)\right),
\end{equation}
where the posterior covariance and mean are, respectively:
\begin{equation}\label{eq:pro:Sigma}
\mathbf\Sigma =\left(\mathbf\Phi^\intercal\mathbf\Phi +\mathbf A\right)^{-1},
\end{equation}
\begin{equation}\label{eq:pro:M}
\mathbf M=\mathbf\Sigma \mathbf\Phi^\intercal\mathbf T.
\end{equation}

In the case of uniform hyperpriors $\bm\upalpha$ and $\mathbf \Omega$, maximising a posteriori \mbox{$p\left(\bm\upalpha,\mathbf\Omega\right|\mathbf T)\propto p\left(\mathbf T|\bm\upalpha,\mathbf\Omega\right)p\left(\bm\upalpha\right)p\left(\mathbf\Omega\right)$} is equivalent to maximising the marginal likelihood $p\left(\mathbf T|\bm\upalpha,\mathbf\Omega\right)$, which is given by:
\begin{equation}\label{eq:pro:p(T|a,O)}
p\left(\mathbf T|\bm\upalpha,\mathbf\Omega\right)=\left(2\pi
\right)^{-\frac{VN} 2}\left|\mathbf\Omega \right|^{-\frac{N} 2}\left|\mathbf I+\mathbf\Phi\mathbf A^{-1}\mathbf\Phi ^\intercal\right|^{-\frac{V}
2}\textnormal{exp}\left(-\frac{1} 2\textnormal{tr}\left(\mathbf\Omega ^{-1}\mathbf T^\intercal\left(\mathbf I+\mathbf\Phi \mathbf A^{-1}\mathbf\Phi^\intercal\right)^{-1}\mathbf T\right)\right).
\end{equation}

\subsection{Marginal likelihood maximisation}
The same method of accelerating the univariate relevance vector machine~\citep{tipping2003fast} is used to accelerate the proposed algorithm.

The log of Eq.~(\ref{eq:pro:p(T|a,O)}) is an objective function:
\begin{equation}\label{eq:pro:L(a,O)}
\mathcal L(\bm\upalpha,\mathbf\Omega)=-\frac{1} 2\left(VN\log(2\pi)+N\log \left|\mathbf\Omega \right|+V\log
\left|\mathbf C\right|+\textnormal{tr}\left(\mathbf\Omega^{-1}\mathbf T^\intercal \mathbf C^{-1} \mathbf T\right)\right),
\end{equation}
where $\mathbf C=\mathbf I+\mathbf\Phi \mathbf A^{-1}\mathbf\Phi ^\intercal\in\mathbb R^{N\times N}$, and by considering the dependence of $\mathcal L(\bm\upalpha,\mathbf\Omega)$ on a single hyperparameter $\alpha_i,i\in\{0,1,\ldots,N\}$, $\mathbf C$ is decomposed as the following two parts:
\begin{equation}\begin{split}
\mathbf C&=\mathbf I+\underset{m\neq i}{\sum }\alpha _m^{-1}\bm\upphi _m\bm\upphi _m^\intercal+\alpha _i^{-1}\bm\upphi _i\bm\upphi _i^\intercal\\
&=\mathbf C_{-i}+\alpha _i^{-1}\bm\upphi _i\bm\upphi _i^\intercal,
\end{split}\end{equation}
where $\mathbf C_{-i}\in\mathbb R^{N\times N}$ is $\mathbf C$ with the contribution of a basis vector $\bm\upphi _i\in\mathbb R^{N\times 1}$ removed, and
\begin{equation}
\bm\upphi_i=\left\{\begin{array}{cl}
[1~1~\ldots~1]^\intercal,&\mbox{ if $i=0$} \\
~[K(\mathbf x_i,\mathbf x_1)~K(\mathbf x_i,\mathbf x_2)~\ldots~K(\mathbf x_i,\mathbf x_N)]^\intercal,&\mbox{ otherwise}
\end{array}\right..
\end{equation}
The determinant and inverse matrix of $\mathbf C$ are, respectively:
\begin{equation}
\left|\mathbf C\right|=\left|\mathbf C_{-i}\right|\left(1+\alpha _i^{-1}\bm\upphi_i^\intercal\mathbf C_{-i}^{-1}\bm\upphi_i\right),
\end{equation}
by Sylvester's determinant theorem, and
\begin{equation}
\mathbf C^{-1}=\mathbf C_{-i}^{-1}-\frac{\mathbf C_{-i}^{-1}\bm\upphi_i\bm\upphi_i^\intercal\mathbf C_{-i}^{-1}}{\alpha _i+\bm\upphi_i^\intercal\mathbf C_{-i}^{-1}\bm\upphi_i},
\end{equation}
by Woodbury matrix identity. From these, $\mathcal L(\bm\upalpha,\mathbf\Omega)$ in Eq.~(\ref{eq:pro:L(a,O)}) can be decomposed into $\mathcal L(\bm\upalpha_{-i},\mathbf\Omega)$, the marginal likelihood with $\bm\upphi_i$ excluded, and $\ell(\alpha_i,\mathbf\Omega)$, the isolated marginal likelihood of $\bm\upphi_i$:
\begin{equation}
\begin{split}
\mathcal L(\bm\upalpha,\mathbf\Omega)=&-\frac{1} 2\left(VN\log(2\pi) +N\log \left|\mathbf\Omega
\right|+V\log\left|\mathbf C_{-i}\right|+\textnormal{tr}\left(\mathbf\Omega ^{-1}\mathbf T^\intercal \mathbf C_{-i}^{-1}\mathbf T\right)\right)\\
&-\frac 1 2\left(-V\log\alpha _i+V\log\left(\alpha_i+\bm\upphi_i^\intercal\mathbf C_{-i}^{-1}\bm\upphi_i\right)-\textnormal{tr}\left(\mathbf\Omega
^{-1}\mathbf T^\intercal\frac{\mathbf C_{-i}^{-1}\bm\upphi_i\bm\upphi_i^\intercal\mathbf C_{-i}^{-1}}{\alpha _i+\bm\upphi_i^\intercal \mathbf C_{-i}^{-1}\bm\upphi_i}\mathbf T\right)\right)\\
=&\mathcal L(\bm\upalpha_{-i},\mathbf\Omega)+\frac 1 2\left(V\log \alpha_i-V\log \left(\alpha
_i+s_i\right)+\frac{\textnormal{tr}\left(\mathbf\Omega ^{-1}\mathbf q_i^\intercal \mathbf q_i\right)}{\alpha _i+s_i}\right)\\
=&\mathcal L(\bm\upalpha_{-i},\mathbf\Omega)+\ell(\alpha_i,\mathbf\Omega),
\end{split}
\end{equation}
where $s_i$ and $\mathbf q_i\in\mathbb R^{1\times V}$ are defined as, respectively:
\begin{subequations}\label{eq:pro:sqOriginal}
	\begin{equation}
	s_i\defeq\bm\upphi_i^\intercal\mathbf C_{-i}^{-1}\bm\upphi_i,
	\end{equation}
	\begin{equation}
	\mathbf q_i\defeq\bm\upphi_i^\intercal\mathbf C_{-i}^{-1}\mathbf T.
	\end{equation}
\end{subequations}

To avoid the matrix inversion of $\mathbf C_{-i}$ in Eq.~(\ref{eq:pro:sqOriginal}), which requires the time complexity of $O\left(N^3\right)$, $s_i'$ and $\mathbf q_i'\in\mathbb R^{1\times V}$ are computed as (by the Woodbury matrix identity):\footnote{$s_i'=\bm\upphi_i^\intercal\bm\upphi_i$ and $\mathbf q_i'=\bm\upphi_i^\intercal \mathbf T$ when $\alpha_i=\infty,\forall i$.}
\begin{subequations}\label{eq:pro:sqPrime}
	\begin{equation}\begin{split}
	s_i'&=\bm\upphi_i^\intercal\mathbf C^{-1}\bm\upphi_i\\
	&=\bm\upphi_i^\intercal\bm\upphi_i-\bm\upphi_i^\intercal\mathbf\Phi \mathbf\Sigma \mathbf\Phi ^\intercal\bm\upphi_i,
	\end{split}\end{equation}
	\begin{equation}\begin{split}
	\mathbf q_i'&=\bm\upphi_i^\intercal\mathbf C^{-1}\mathbf T\\
	&=\bm\upphi_i^\intercal \mathbf T-\bm\upphi_i^\intercal\mathbf\Phi \mathbf\Sigma \mathbf\Phi^\intercal\mathbf T,
	\end{split}\end{equation}
\end{subequations}
and then $s_i$ and $\mathbf q_i$ in Eq.~(\ref{eq:pro:sqOriginal}) are computed as:\footnote{$s_i=s_i'$ and $\mathbf q_i=\mathbf q_i'$ when $\alpha_i=\infty$.}
\begin{subequations}\label{eq:pro:sq}
	\begin{equation}
	s_i=\frac{\alpha_is_i'}{\alpha_i-s_i'},
	\end{equation}
	\begin{equation}
	\mathbf q_i=\frac{\alpha_i\mathbf q_i'}{\alpha_i-s_i'}.
	\end{equation}
\end{subequations}

$\displaystyle\frac{\partial\ell(\alpha_i,\mathbf\Omega)}{\partial\alpha_i}=0$ leads to that $\mathcal L(\bm\upalpha,\mathbf\Omega)$ has a unique maximum with respect to $\alpha_i$ when:
\begin{equation}\label{eq:pro:alpha}
\alpha_i=\left\{\begin{array}{cl}
\displaystyle\frac{s_i^2}{\displaystyle\frac{\textnormal{tr}\left(\mathbf\Omega^{-1}\mathbf q_i^\intercal\mathbf q_i\right)} V-s_i},&\textnormal{ if }\displaystyle\frac{\textnormal{tr}\left(\mathbf\Omega ^{-1}\mathbf q_i^\intercal \mathbf q_i\right)}V>s_i\\
~\infty,&\textnormal{ if }\displaystyle\frac{\textnormal{tr}\left(\mathbf\Omega ^{-1}\mathbf q_i^\intercal \mathbf q_i\right)}V\le s_i
\end{array}\right.,
\end{equation}
which implies that:
\begin{itemize}
  \item If $\bm\upphi_i$ is ``in the model" (i.e. $\alpha_i<\infty$) and $\displaystyle\frac{\textnormal{tr}\left(\mathbf\Omega ^{-1}\mathbf q_i^\intercal \mathbf q_i\right)}V> s_i$; then, $\alpha_i$ is re-estimated,
  \item If $\bm\upphi_i$ is ``in the model" (i.e. $\alpha_i<\infty$) yet $\displaystyle\frac{\textnormal{tr}\left(\mathbf\Omega ^{-1}\mathbf q_i^\intercal \mathbf q_i\right)}V\le s_i$; then, $\bm\upphi_i$ may be deleted (i.e. $\alpha_i$ is set to be $\infty$),
  \item If $\bm\upphi_i$ is ``out of the model" (i.e.~$\alpha_i=\infty$) yet $\displaystyle\frac{\textnormal{tr}\left(\mathbf\Omega ^{-1}\mathbf q_i^\intercal \mathbf q_i\right)}V>s_i$; then, $\bm\upphi_i$ may be added (i.e.~$\alpha_i$ is set to be a finite value).
\end{itemize}

In addition, $\displaystyle\frac{\partial\mathcal L(\bm\upalpha,\mathbf\Omega)}{\partial\mathbf\Omega}=\mathbf0$, where $\mathbf0$ is a $V\times V$ zero matrix, leads to that $\mathcal L(\bm\upalpha,\mathbf\Omega)$ has a unique maximum with respect to $\mathbf\Omega$ when:
\begin{equation}\label{eq:pro:Omega}
\mathbf\Omega=\dfrac{\mathbf T^\intercal(\mathbf T-\mathbf\Phi \mathbf M)}{N}.
\end{equation}
%However, $\mathbf T^\intercal(\mathbf T-\mathbf\Phi \mathbf M)$ is not the equation form of the sample covariance matrix $(\mathbf T-\mathbf\Phi \mathbf M)^\intercal(\mathbf T-\mathbf\Phi \mathbf M)$ and it can be inferred that the assumption $\mathbf\Omega=\frac{\mathbb E\left[\mathbf E^\intercal \mathbf E\right]}{N}=\frac{\mathbb E\left[\mathbf W^\intercal \mathbf W\right]}{\textnormal{tr}\left(\mathbf A^{-1}\right)}$ causes Eq.~(\ref{eq:pro:Omega}). In other words, the assumption that $\mathbf\Omega$ implies the among-column covariance matrix of the weight $\mathbf W$ as well as that of the error $\mathbf E$ induces the slightly different form of the sample covariance matrix. Therefore, the sample covariance matrix is also experimented:
%\begin{equation}\label{eq:smplCov}
%\mathbf\Omega=\dfrac{(\mathbf T-\mathbf\Phi \mathbf M)^\intercal(\mathbf T-\mathbf\Phi \mathbf M)}{N-1},
%\end{equation}
%and Eq.~(\ref{eq:pro:Omega}) and Eq.~(\ref{eq:smplCov}) are compared in Section~\ref{sec:exp}.

\subsection{Expectation-maximisation (EM) algorithm}
\begin{algorithm}
\caption{Proposed EM algorithm of MRVR.}\label{alg:pro}
\KwIn{$\mathbf T\in\mathbb R^{N\times V}$ and $\bm\upphi_i\in\mathbb R^{N\times 1},\forall i=\{0,1,\ldots,N\}$, where $N$ is the number of training samples, and $V$ is the number of output dimensions}
\KwOut{$\mathbf\Sigma\in\mathbb R^{M\times M}$, $\mathbf M\in\mathbb R^{M\times V}$, and $\mathbf\Omega\in\mathbb R^{V\times V}$, where $M$ is the number of basis functions in the model}
\tcp{Initilisation}
$\alpha_i\gets\infty,\forall i=\{0,1,\ldots,N\}$\\
$\bar{\mathbf t}\gets\frac{1}{N}\sum_{i=1}^{N}\mathbf t_i$, where $\bar{\mathbf t}\in\mathbb{R}^{1\times V}$, and $\mathbf t_i\in\mathbb{R}^{1\times V}$ is the $i$-th row vector of $\mathbf T$.\\
$\mathbf\Omega\gets\frac{0.1}{N-1}\sum_{i=1}^{N}(\mathbf t_i-\bar{\mathbf t})^\intercal(\mathbf t_i-\bar{\mathbf t})$, where $\mathbf\Omega\in\mathbb{R}^{V\times V}$ is a covariance matrix.\\
%Initialise with a single basis vector $\bm\upphi_i$\footnotemark[\ref{fnt:init}] using Eq.~(\ref{eq:pro:sqOriginal}) and Eq.~(\ref{eq:pro:alpha}):
%$\alpha_i\gets\frac{\left\|\bm\upphi_i\right\|^2}{\frac{\textnormal{tr}\left(\mathbf\Omega ^{-1}\mathbf T^\intercal\bm\upphi_i\bm\upphi_i^\intercal \mathbf T\right)}{V\left\|\bm\upphi_i\right\|^2}-1}$,
%and all other $\alpha_i$ are set to be $\infty$.\\
convergence$\gets$false\\
$n\gets1$, where $n$ is the iteration number\\
$M\gets0$, where $M$ is the number of basis functions.\\
\While{convergence=false}{
	\tcp{maximisation step}
	\For{$i\gets0~\text{to}~N$}{
		Update $s_i'$ and $\mathbf q_i'$ using Eq.~(\ref{eq:pro:sqPrime}).\\
		Update $s_i$ and $\mathbf q_i$ using Eq.~(\ref{eq:pro:sq}).\\
		$\theta_i\gets\frac{\textnormal{tr}\left(\mathbf\Omega ^{-1}\mathbf q_i^\intercal \mathbf q_i\right)}V-s_i$\\
		\uIf{$\theta_i>0$}{
			\eIf($z_i\gets$`re-estimation'){$\alpha_i<\infty$}{
				$\tilde\alpha_i\gets\frac{s_i^2}{\theta_i}$\\
				Update $2\Delta\mathcal L_i$ using Eq.~(\ref{eq:pro:2DLest}).}
			($z_i\gets$`addition'){
				Update $2\Delta\mathcal L_i$ using Eq.~(\ref{eq:pro:2DLadd}).}
		}
		\uElseIf($z_i\gets$`deletion'){$\alpha_i<\infty$}{
			Update $2\Delta\mathcal L_i$ using Eq.~(\ref{eq:pro:2DLdel}).
		}
		\Else{$2\Delta\mathcal L_i\gets-\infty$}
	}
	$i\gets\arg\max_i2\Delta\mathcal L_i$~\tcp{Select $i$ which gives the greatest increase of the marginal likelihood}
	\Switch{$z_i$}{
		\uCase{`re-estimation'}{
			$\Delta\log\alpha\gets\log\frac{\alpha_i}{\tilde\alpha_i}$\\
			$\alpha_i\gets\tilde\alpha_i$\\
			\tcp{Check convergence (convergence criteria are the same as those in~\citep{tipping2003fast})}
			\If{$|\Delta\log\alpha|<0.1$}
				{convergence$\gets$true\\
				\For{$i\gets0~\text{to}~N$}{
					\If(\tcp*[h]{if $\bm{\upphi}_i$ is "out of the model"}){$\alpha_i=\infty$}{
						\If(\tcp*[h]{if $\bm{\upphi}_i$ may be added}){$\theta_i>0$}{convergence$\gets$false\\break for loop}
					}
				}
			}
	    }
	    \uCase{`addition'}{
	    	$\alpha_i\gets\frac{s_i^2}{\theta_i}$\\
	    	$M\gets M+1$
	    }
	    \uCase{`deletion'}{
	    	$\alpha_i\gets\infty$\\
	    	$M\gets M-1$
	    }
	}
	\If{$n\neq1$}{
		Update $\mathbf\Omega$ using Eq.~(\ref{eq:pro:Omega}).\\ %or Eq.~(\ref{eq:smplCov}).\\
	}	
	
	\tcp{Expectation step}
	Sequentially update \rmnum1) $\mathbf\Phi\in\mathbb R^{N\times M}$, $\mathbf A\in\mathbb R^{M\times M}$, \rmnum2) $\mathbf\Sigma\in\mathbb R^{M\times M}$, and \rmnum3)~$\mathbf M\in\mathbb R^{M\times V}$ using Eq.~(\ref{eq:pro:Sigma}) and Eq.~(\ref{eq:pro:M}), where $\mathbf\Phi,~\mathbf\Sigma$, and $\mathbf M$ contain only $M$ basis functions that are currently included in the model, and the diagonal matrix $\mathbf A$ consists of $M$ hyperparameters of $\alpha_i$ that are currently included in the model.\\
	$n\gets n+1$\\
}
\end{algorithm}
%\stepcounter{footnote}\footnotetext{$\bm{\upphi}_i$ which gives the greatest initial marginal likelihood using Eq.~(\ref{eq:pro:2DLadd}) is selected.\label{fnt:init}}
%\stepcounter{footnote}\footnotetext{The convergence criteria are the same as~\cite{tipping2003fast}.\label{fnt:cnvg}}

Algorithm~\ref{alg:pro}, an EM algorithm  to maximise the marginal likelihood, starts without any basis vector (i.e. $M=0$) and selects the basis vector $\bm\upphi_i$ which gives the maximum change of the marginal likelihood $\mathcal L(\bm\upalpha,\mathbf\Omega)$ of Eq.~(\ref{eq:pro:L(a,O)}) at every iteration.

For efficient computation of the EM algorithm, quantities $\mathbf\Phi\in\mathbb R^{N\times M}$ and $\mathbf\Sigma\in\mathbb R^{M\times M}$ contain only $M$ ($M\le N+1$ is always satisfied) basis functions that are currently included in the model (i.e. $\bm\upphi_i$ which satisfies $\alpha_i<\infty$), and the diagonal matrix $\mathbf A$ consists of $M$ hyperparameters of $\alpha_i$ that are currently included in the model (i.e. $\alpha_i$ which satisfies $\alpha_i<\infty$).

From Eq.~(\ref{eq:pro:L(a,O)}), the change in the marginal likelihood can be written as
\begin{equation}\begin{split}\label{eq:pro:2DL}
2\Delta\mathcal L&=2\left(\mathcal L(\tilde{\bm{\upalpha}},\mathbf\Omega)-\mathcal L(\bm\upalpha,\mathbf\Omega)\right)\\
&=V\log\frac{\left|\mathbf C\right|}{\left|\tilde{\mathbf C}\right|}+\textnormal{tr}\left(\mathbf\Omega
^{-1}\mathbf T^\intercal\left(\mathbf C^{-1}-\tilde{\mathbf C}^{-1}\right)\mathbf T\right),
\end{split}\end{equation}
where updated quantities are denoted by a tilde (e.g., $\tilde{\bm{\upalpha}}$ and $\tilde{\mathbf C}$). Eq.~(\ref{eq:pro:2DL}) differs
according to whether $\alpha_i$ is re-estimated, added, or deleted:
\paragraph{Re-estimation} as $\mathbf C=\mathbf C_{-i}+\alpha _i^{-1}\bm\upphi_i\bm\upphi_i^\intercal$ and $\tilde{\mathbf C}=\mathbf C_{-i}+\tilde\alpha_i^{-1}\bm\upphi_i\bm\upphi_i^\intercal$,
\begin{equation}\label{eq:pro:2DLest}
2\Delta\mathcal L_i=\frac{\textnormal{tr}\left(\mathbf\Omega^{-1} \mathbf q_i'^\intercal \mathbf q'_i\right)}{s_i'+\left(\tilde\alpha_i^{-1}-\alpha
			_i^{-1}\right)^{-1}}-V\log \left(1+s_i'\left(\tilde\alpha_i^{-1}-\alpha
			_i^{-1}\right)\right),
\end{equation}
where $\tilde\alpha_i$ is re-estimated $\alpha_i$,
\paragraph{Addition} as $\mathbf C=\mathbf C_{-i}$ and $\tilde{\mathbf C}=\mathbf C_{-i}+\tilde\alpha_i^{-1}\bm\upphi_i\bm\upphi_i^\intercal$, where $\tilde\alpha_i=\displaystyle\frac{s_i^2}{\displaystyle\frac{\textnormal{tr}\left(\mathbf\Omega^{-1}\mathbf q_i^\intercal\mathbf q_i\right)} V-s_i}$,
\begin{equation}\label{eq:pro:2DLadd}
2\Delta\mathcal L_i=\frac{\textnormal{tr}\left(\mathbf\Omega^{-1} \mathbf q_i'^\intercal \mathbf q'_i\right)-Vs_i'}{s_i'}+V\log\frac{Vs_i'}{\textnormal{tr}\left(\mathbf\Omega^{-1} \mathbf q_i'^\intercal \mathbf q'_i\right)},
\end{equation}
\paragraph{Deletion} as $\mathbf C=\mathbf C_{-i}+\alpha _i^{-1}\bm\upphi_i\bm\upphi_i^\intercal$ and $\tilde{\mathbf C}=\mathbf C_{-i}$,
\begin{equation}\label{eq:pro:2DLdel}
2\Delta\mathcal L_i=\frac{\textnormal{tr}\left(\mathbf\Omega^{-1} \mathbf q_i'^\intercal \mathbf q'_i\right)}{s_i'-\alpha_i}-V\log\left(1-\frac{s_i'}{\alpha_i}\right).
\end{equation}

\subsection{Making predictions}
We can predict both a mean vector $\mathbf y_*\in\mathbb R^{1\times V}$ and a covariance matrix $\mathbf\Omega_*\in\mathbb R^{V\times V}$ from a new input vector $\mathbf x_*\in\mathbb R^{U\times 1}$ based on
both \rmnum1) Eq.~(\ref{eq:pro:T}), the model specification, and \rmnum2) Eq.~(\ref{eq:pro:p(W|T,a,O)}), the posterior distribution over the weights, conditioned on the most probable (MP) hyperparameters: $\bm{\upalpha}_\textnormal{MP}\in\mathbb R_{>0}^{M\times 1}$ and $\mathbf\Omega_\textnormal{MP}\in\mathbb R^{V\times V}$, obtained from Algorithm~\ref{alg:pro}. Predictive distribution of $\mathbf t_*$ is jointly normally distributed as
\begin{equation}
p(\mathbf t_*|\mathbf T,\bm{\upalpha}_\textnormal{MP},\mathbf\Omega_\textnormal{MP})=\mathcal N(\mathbf t_*|\mathbf y_*,\mathbf\Omega_*),
\end{equation}
with
\begin{equation}
\mathbf y_*=\bm\upphi(\mathbf x_*)^\intercal\mathbf M,
\end{equation}
and
\begin{equation}\label{eq:pro:Omega_*}
\mathbf\Omega_*=\mathbf\Omega_\textnormal{MP}\left(1+\bm\upphi(\mathbf x_*)^\intercal\mathbf\Sigma\bm\upphi(\mathbf x_*)\right),\footnote{Eq.~(\ref{eq:pro:Omega_*}) is obtained by the property that the covariance between two elements $W_{i,j}$ and $W_{i',j'}$ is the covariance between the rows $i$ and $i'$, i.e. $\mathbf\Sigma$, multiplied by the covariance between the columns $j$ and $j'$, i.e. $\mathbf\Omega_\textnormal{MP}$~\cite[p.~311]{arnold1981theory}.}
\end{equation}
where $\bm\upphi(\mathbf x)\in\mathbb R^{M\times 1}$ comes from only $M$ basis functions that are included in the model after the EM algorithm. The predictive covariance matrix $\mathbf\Omega_*$ comprises the two components: the estimated noise on the training data $\mathbf\Omega_\textnormal{MP}$ and that due to the uncertainty in the prediction of the weights $\mathbf\Omega_\textnormal{MP}\bm\upphi(\mathbf x_*)^\intercal\mathbf\Sigma\bm\upphi(\mathbf x_*)$, where $\bm\upphi(\mathbf x_*)^\intercal\mathbf\Sigma\bm\upphi(\mathbf x_*)\in\mathbb R_{\ge0}$ by the fact that a covariance matrix is always positive semidefinite. They share $\mathbf\Omega_\textnormal{MP}$ by the assumption of $\mathbf\Omega=\frac{\mathbb E\left[\mathbf E^\intercal \mathbf E\right]}{N}=\frac{\mathbb E\left[\mathbf W^\intercal \mathbf W\right]}{\textnormal{tr}\left(\mathbf A^{-1}\right)}$ in Section~\ref{sec:pro:model}.

\subsection{Algorithm complexity}
Matrix inversion of $\mathbf{\Omega}\in\mathbb R^{V\times V}$ in Eq.~(\ref{eq:pro:alpha}) and that of the $M\times M$ matrix in Eq.~(\ref{eq:pro:Sigma}) determine \rmnum1) the time complexity of the proposed algorithm as \mbox{$O\left(V^3+M^3\right)$} and \rmnum2) the memory complexity as $O\left(V^2+M^2\right)$, where $V$ is the number of output dimensions, and $M$ is the number of basis functions.\footnote{The matrix multiplication to calculate $s_i'$ and $\mathbf q_i'$ in Eq.~\eqref{eq:pro:sqPrime} for all $i\in\{1,2,\ldots,N\}$ at the 9-th line of Algorithm~\ref{alg:pro} does not influence the time complexity because the matrix multiplication $\mathbf\Phi \mathbf\Sigma\mathbf\Phi ^\intercal$ is pre-calculated. In other words, the time complexity of the matrix multiplication is $O\left(M^3\right)$, not $O\left(NM^3\right)$, because $\mathbf\Phi \mathbf\Sigma\mathbf\Phi ^\intercal$ is independent of $i$.}

\section{Experimental results}\label{sec:results}
\subsection{An example of MRVR}
\begin{figure}
    \centering
    \begin{subfigure}[t]{0.5\textwidth}
            \includegraphics[width=\textwidth]{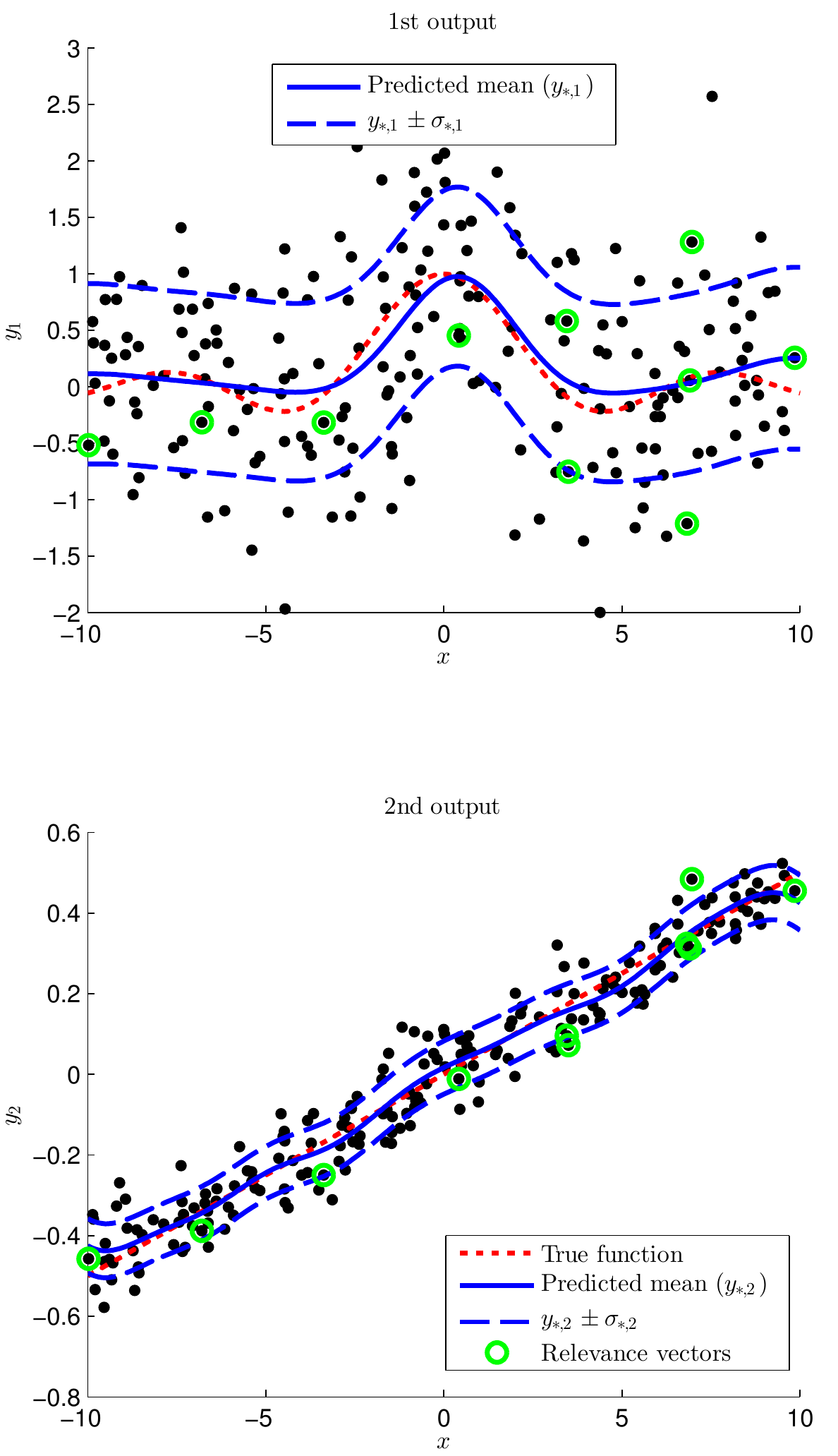}
            \caption{Existing method ($\sigma_{*,j}^2$ is equal to Eq.~(\ref{eq:exi:sigma_*^2}))}
            \label{fig:ex:a}
    \end{subfigure}%
    ~ %add desired spacing between images, e. g. ~, \quad, \qquad, \hfill etc.
    \begin{subfigure}[t]{0.5\textwidth}
            \includegraphics[width=\textwidth]{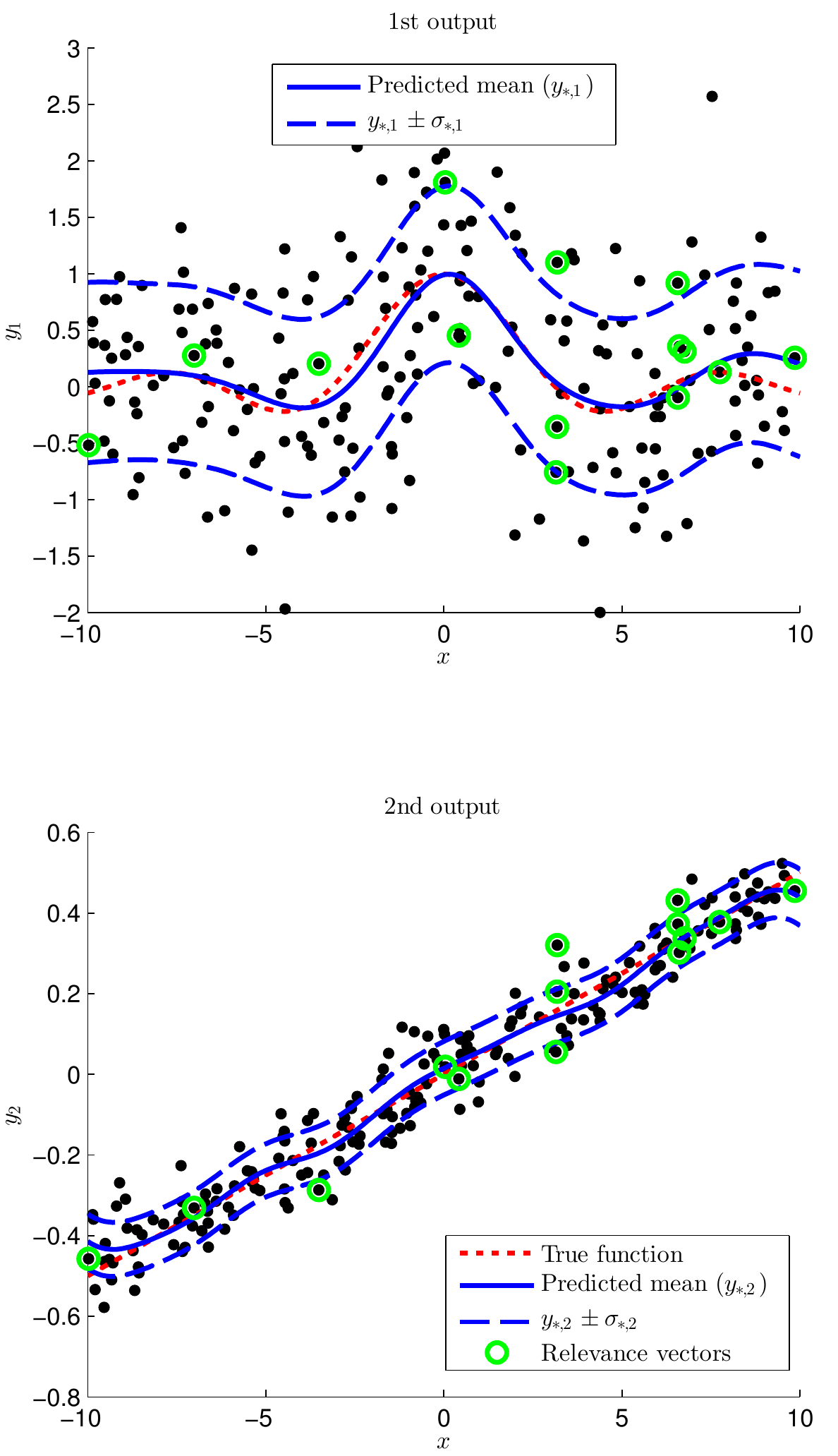}
            \caption{Proposed method ($\sigma_{*,j}^2$ is equal to the $j$-th diagonal element of $\mathbf\Omega_*$ in Eq.~(\ref{eq:pro:Omega_*}))}
            \label{fig:ex:b}
    \end{subfigure}
    \caption{An example of MRVR (when $U=1,V=2,N=200$, and the Gaussian kernel $K\left(\mathbf x,\mathbf x'\right)=\exp\left(-\frac{\|\mathbf x-\mathbf x'\|^2}{2\lambda^2}\right)$ with a free parameter $\lambda=1.6$ is used)}
    \label{fig:ex}
\end{figure}

%the thin plate spline kernel $K(\mathbf x,\mathbf x')=\frac{\|\mathbf x-\mathbf x'\|^2}{\lambda^2}\log\frac{\|\mathbf x-\mathbf x'\|}{\lambda}$

Fig.~\ref{fig:ex} shows an example of the MRVR results obtained using the two methods when the true function of each output dimension is the sinc function and the linear function, respectively. Fig.~\ref{fig:ex:a} and Fig.~\ref{fig:ex:b} show slightly different results although the same training samples are used. %but the predicted standard deviation depends on $x$ in common because of the common term $\bm\upphi(\mathbf x)^\intercal\mathbf\Sigma\bm\upphi(\mathbf x)$ in Eq.~(\ref{eq:exi:sigma_*^2}) and Eq.~(\ref{eq:pro:Omega_*}). This means the training samples in Fig.~\ref{fig:ex} are scattered according to a Gaussian distribution (most samples are located between $x=[-10,10]$). Therefore, a higher uncertainty (quantified by the predicted standard deviation $\sigma_{*,j}$) is observed in the rest region.

\subsection{Comparisons of the performance}
The two methods are compared in terms of \rmnum1)~running time (computation time in INTEL\textsuperscript{\textregistered} Core\textsuperscript{TM} i5-3470 CPU and MATLAB\textsuperscript{\textregistered} R2013b), \rmnum2)~the estimation accuracy of the noise covariance matrix, \rmnum3) root-mean-square error (RMSE) between true functions and predicted mean values, and \rmnum4) the number of relevance vectors (RVs), where RVs are those training vectors associated with the remaining non-zero weights (i.e. the number of basis functions $M$ is equal to the number of RVs).

To measure the estimation accuracy of the noise covariance matrix $\mathbf\Omega$, entropy loss $L_1\left(\mathbf\Omega,\hat{\mathbf\Omega}\right)$ and quadratic loss $L_2\left(\mathbf\Omega,\hat{\mathbf\Omega}\right)$ are used (each of these is 0 when $\hat{\mathbf\Omega}=\mathbf\Omega$ and is positive when $\hat{\mathbf\Omega}\neq\mathbf\Omega$)~\cite[pp.~273\textendash274]{anderson1984introduction}:
\begin{equation}
L_1\left(\mathbf\Omega,\hat{\mathbf\Omega}\right)=\textnormal{tr}\left(\hat{\mathbf\Omega}\mathbf\Omega^{-1}\right)-\log\left|\hat{\mathbf\Omega}\mathbf\Omega^{-1}\right|-V,
\end{equation}
\begin{equation}
L_2\left(\mathbf\Omega,\hat{\mathbf\Omega}\right)=\textnormal{tr}\left(\left(\hat{\mathbf\Omega}\mathbf\Omega^{-1}-\mathbf I\right)^2\right),
\end{equation}
where the estimated $V\times V$ covariance matrix of the noise $\hat{\mathbf\Omega}$ is $\mathbf\Omega_\textnormal{MP}$ in the case of the proposed method. In the case of the existing method, $\hat{\mathbf\Omega}$ can be obtained using both \rmnum1) the estimated standard deviation of the noise $\hat{\mathbf D}=\textnormal{diag}(\sigma_{\textnormal{MP},1},\sigma_{\textnormal{MP},2},\ldots,\sigma_{\textnormal{MP},V})$ in Section~\ref{sec:exi:predict} and \rmnum2) the estimated correlation matrix of the noise $\hat{\mathbf R}$:
\begin{equation}
\hat{\mathbf\Omega}=\hat{\mathbf D}\hat{\mathbf R}\hat{\mathbf D},
\end{equation}
where $\hat{\mathbf R}=\tilde{\mathbf D}^{-1}\tilde{\mathbf\Omega}\tilde{\mathbf D}^{-1}$, $\tilde{\mathbf\Omega}=\displaystyle\frac{(\mathbf T-\mathbf\Phi \tilde{\mathbf M})^\intercal(\mathbf T-\mathbf\Phi \tilde{\mathbf M})}{N-1}$, $\tilde{\mathbf M}=[\bm\upmu_1~\bm\upmu_2~\ldots~\bm\upmu_V]\in\mathbb R^{M\times V}$, and \mbox{$\tilde{\mathbf D}=\sqrt{\textnormal{diag}(\tilde{\mathbf\Omega})}$}.

\begin{figure}
	\centering
    \includegraphics[width=.6\textwidth]{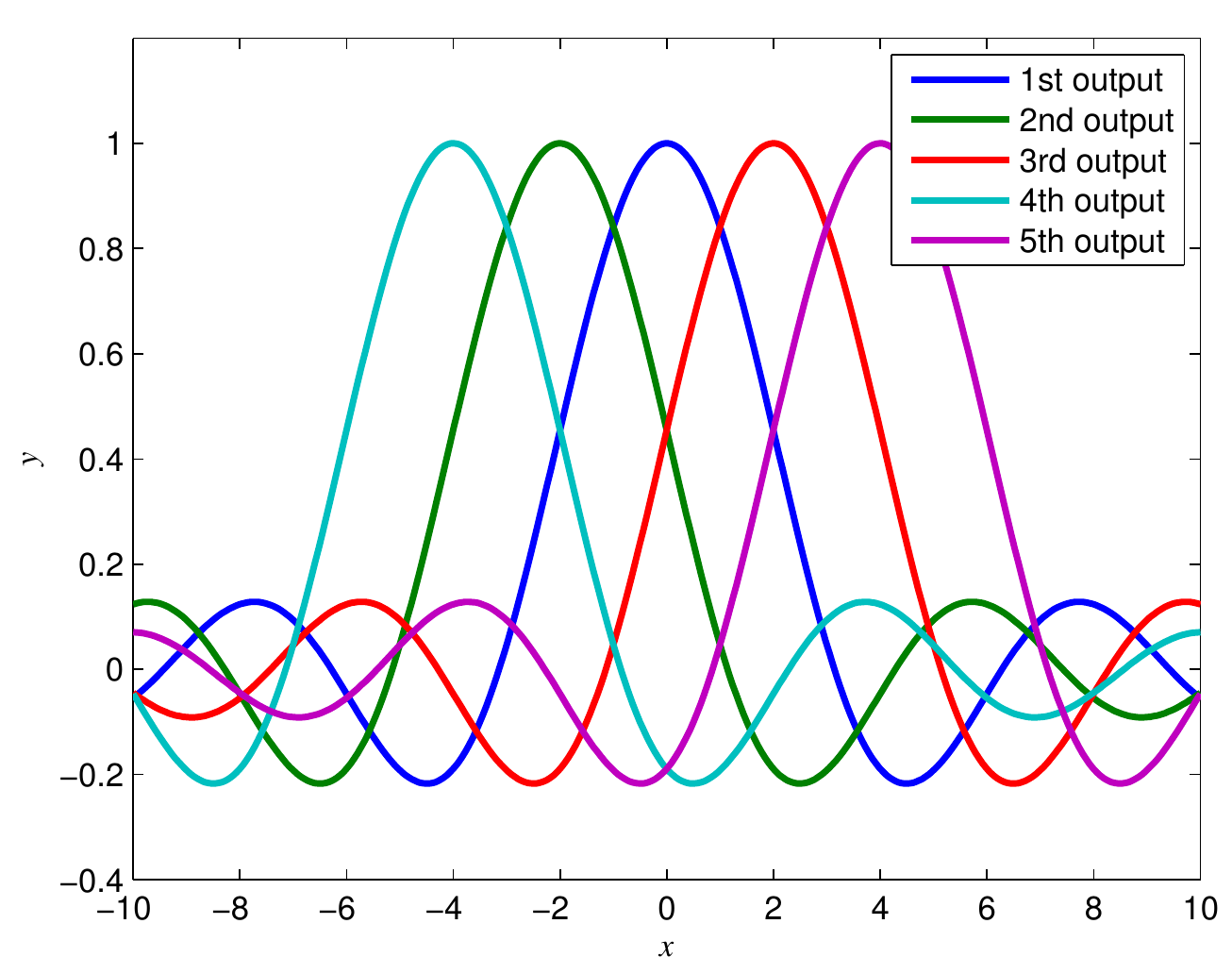}
    \caption{True functions of MC simulations}
    \label{fig:sinc}
\end{figure}

Monte Carlo (MC) simulations with random covariance matrices of the noise were conducted for the performance comparisons. The noise from the random covariance matrix is added to the true functions in Fig.~\ref{fig:sinc} (each output has a sinc function with a translation in the x-axis), and the two methods of MRVR with the Gaussian kernel were performed using the same training samples for a fair comparison.

\begin{table}
\caption{The number of rejections of the null hypothesis of the Jarque\textendash Bera test}
\label{tab:accRate}
\begin{tabularx}{\textwidth}{@{}cccccc@{}}
\hline
\multicolumn{1}{l}{} & \multicolumn{1}{X}{Running time} & \multicolumn{1}{X}{Entropy loss} & \multicolumn{1}{X}{Quadratic loss} & \multicolumn{1}{c}{RMSE} & \multicolumn{1}{X}{The number of RVs} \\ \hline
Existing method & 30 & 30 & 30 & 4 & 16 \\
Proposed method & 30 & 28 & 29 & 3 & 14 \\ \hline
\end{tabularx}
\end{table}

Unpaired two-sample \textit{t}-tests may be used to compare the two methods to determine whether the performance difference is fundamental or whether it is due to random fluctuations~\cite[pp.~631\textendash635]{simon2013evolutionary}, but the normality assumption of the performance measures (i.e. running time, entropy loss, quadratic loss, RMSE, and the number of RVs) of the two methods must be checked. The Jarque\textendash Bera tests $JB=\frac{n}{6}\left(S^2+\frac{1}{4}(K-3)^2\right)$ with the number of observations $n=101$ and a 5\% significance level for 30 cases ($V=\{1,2,3,4,5\}$ and $N=\{50,100,150,200,250,300\}$) were conducted, in which the null hypothesis was that the data of the performance measures came from a normal distribution. Table~\ref{tab:accRate} shows the number of rejections of the null hypothesis. Consequently, the \textit{t}-test can yield misleading results in the case that the null hypothesis is rejected.

Instead of the \textit{t}-test, two-sided Wilcoxon rank sum tests, whose null hypothesis is that two populations have equal median values, were used for the comparisons as they have greater efficiency than the \textit{t}-test on non-normal distributions and are nearly as efficient as the \textit{t}-test on normal distributions~\cite[Chapter~10]{montgomery2013applied}.

\begin{figure}
    \begin{subfigure}[t]{0.5\textwidth}
            \includegraphics[width=\textwidth]{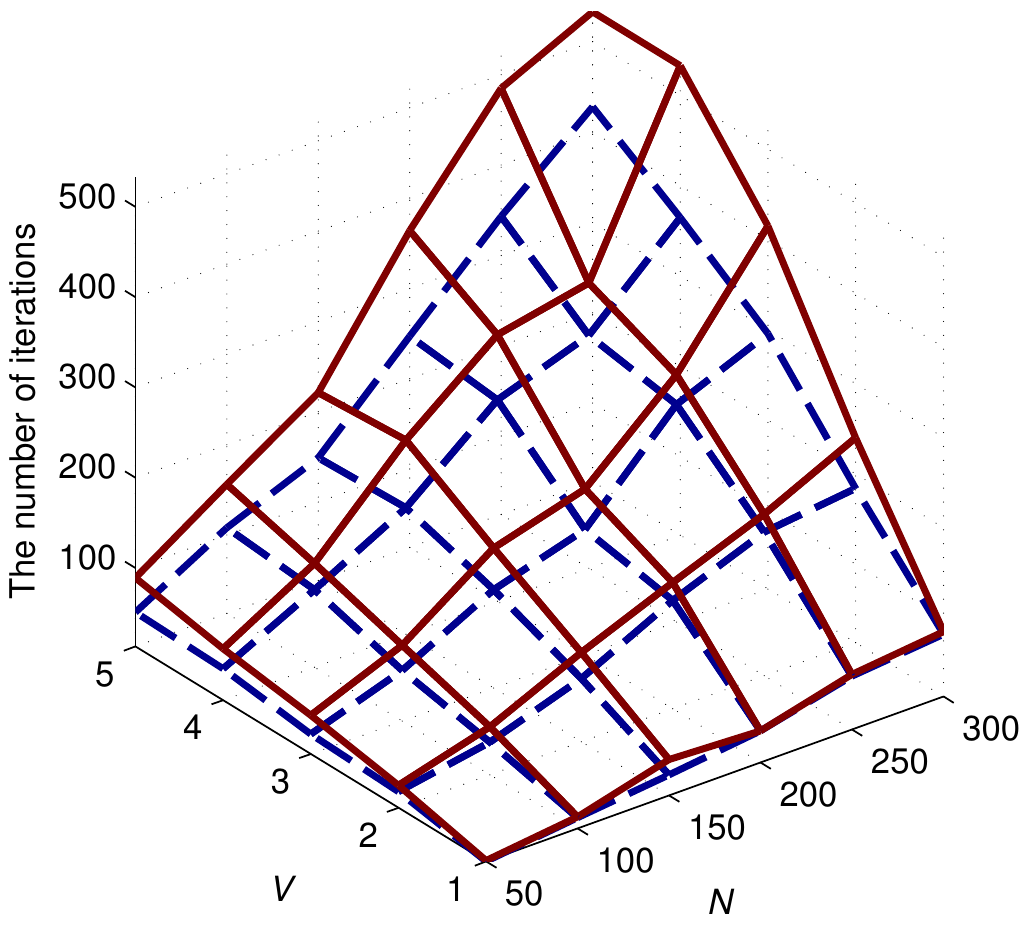}
            \caption{The number of iterations}
            \label{fig:mesh:a}
    \end{subfigure}~    
    \begin{subfigure}[t]{0.5\textwidth}
            \includegraphics[width=\textwidth]{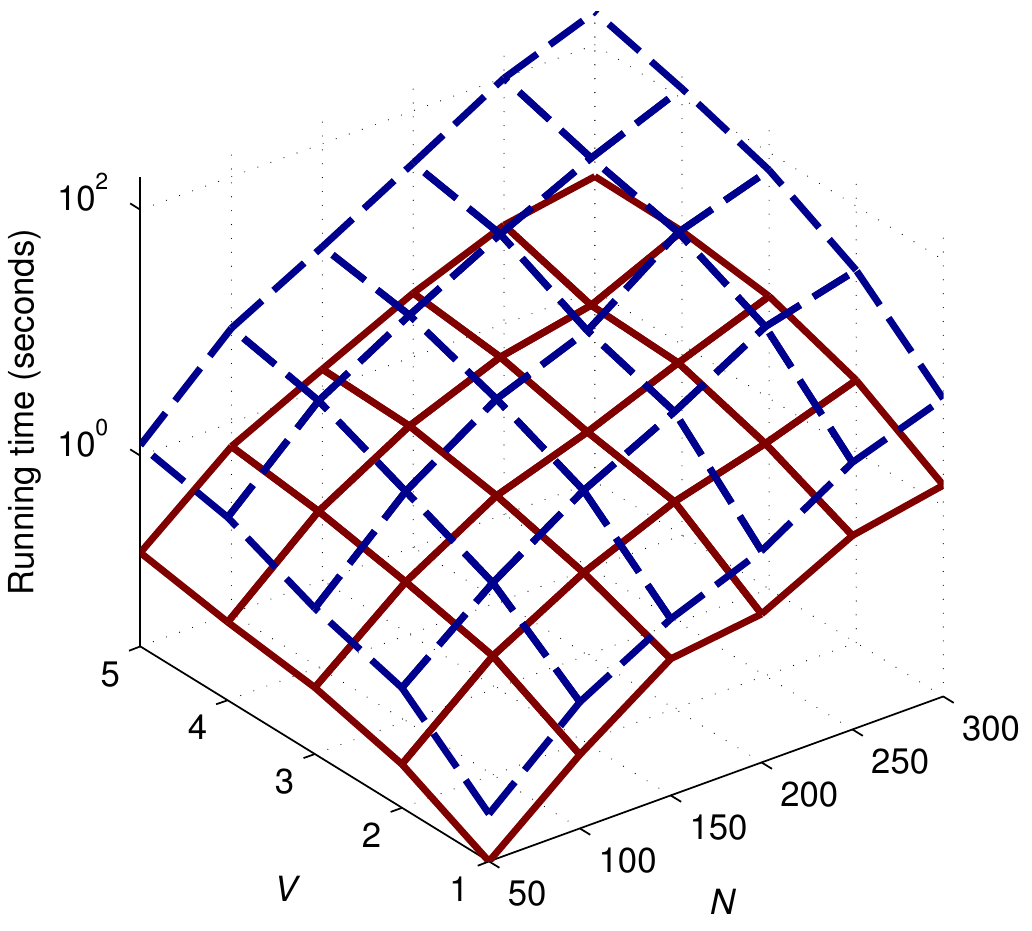}
            \caption{Running time}
            \label{fig:mesh:b}
    \end{subfigure}
    \begin{subfigure}[t]{0.5\textwidth}
            \includegraphics[width=\textwidth]{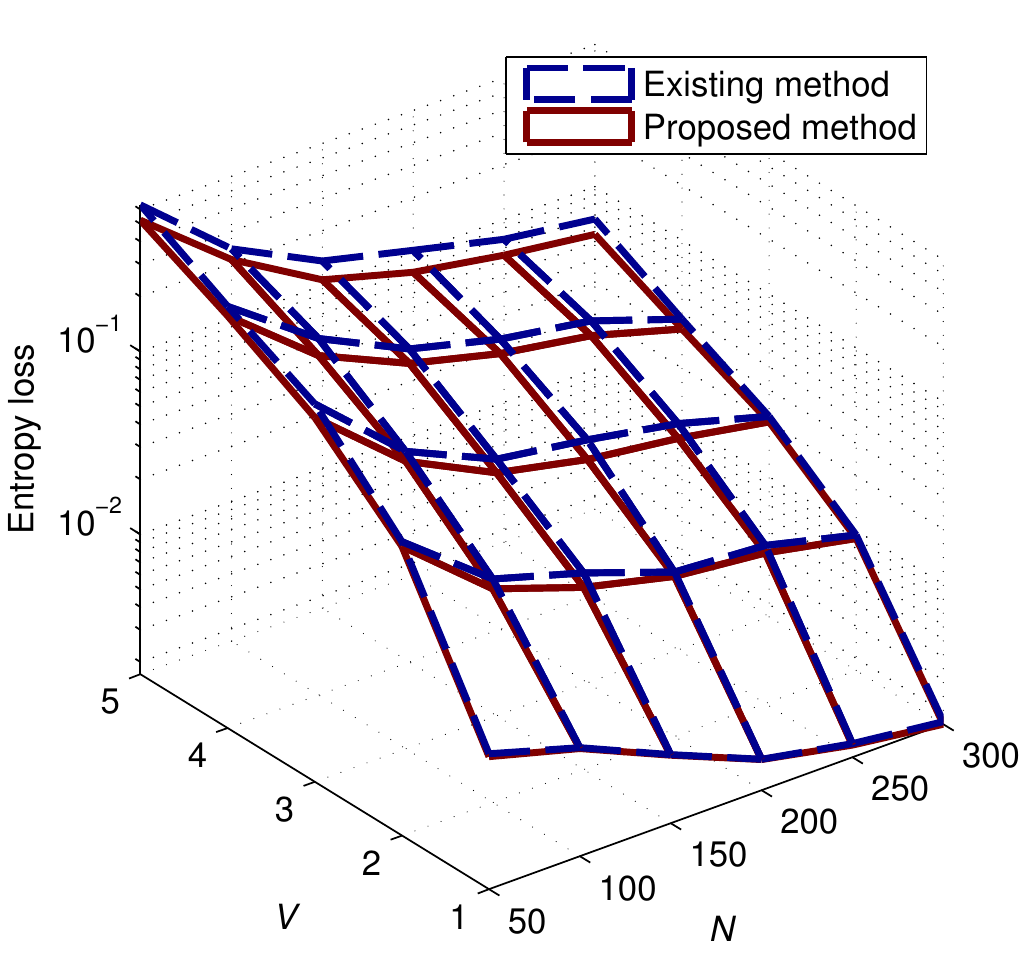}
            \caption{Entropy loss}
            \label{fig:mesh:c}
    \end{subfigure}~
    \begin{subfigure}[t]{0.5\textwidth}
            \includegraphics[width=\textwidth]{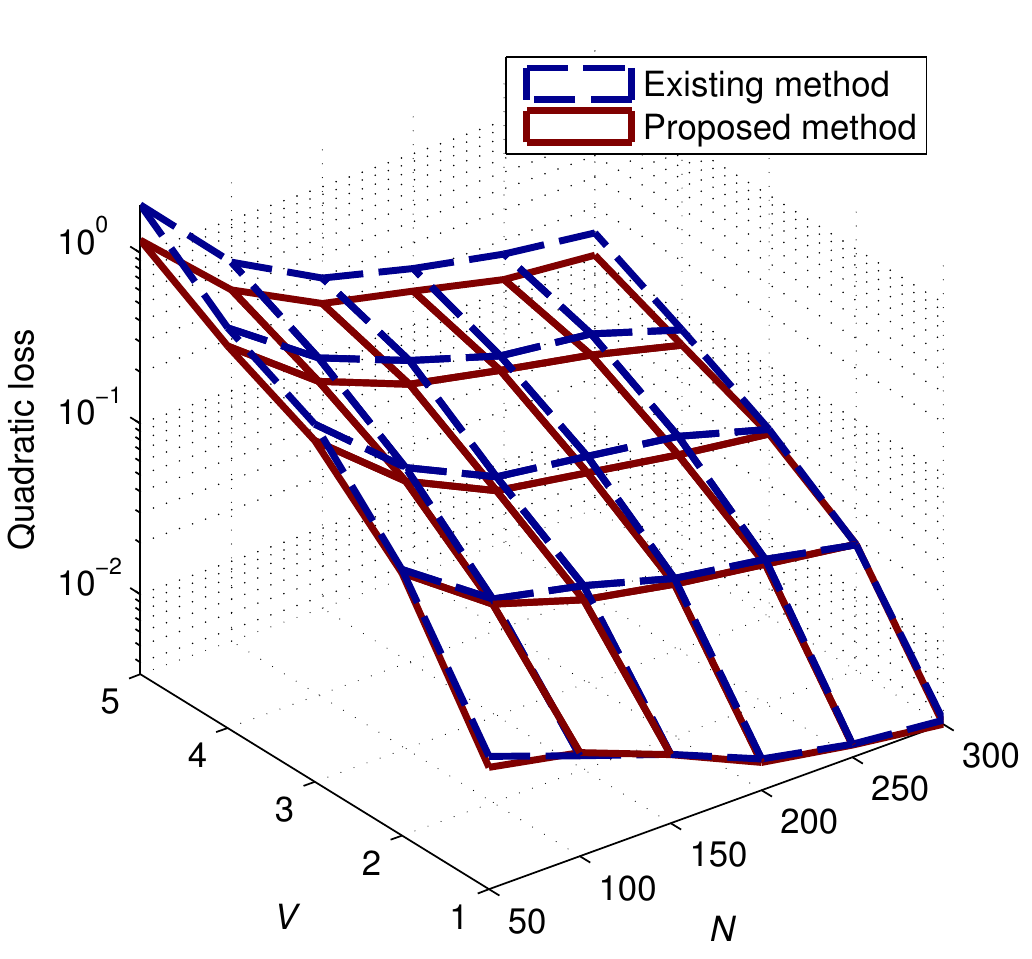}
            \caption{Quadratic loss}
            \label{fig:mesh:d}
    \end{subfigure}
    \begin{subfigure}[t]{0.5\textwidth}
            \includegraphics[width=\textwidth]{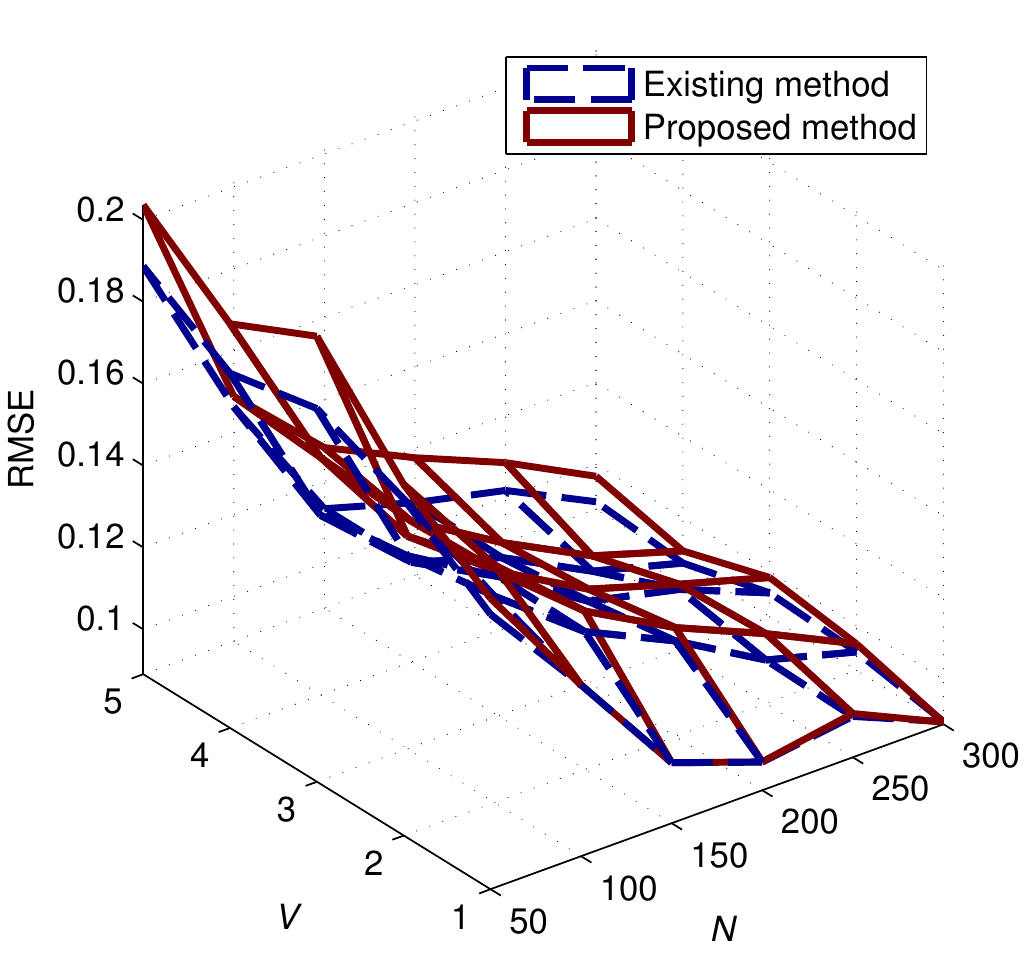}
            \caption{RMSE}
            \label{fig:mesh:e}
    \end{subfigure}~
    \begin{subfigure}[t]{0.5\textwidth}
            \includegraphics[width=\textwidth]{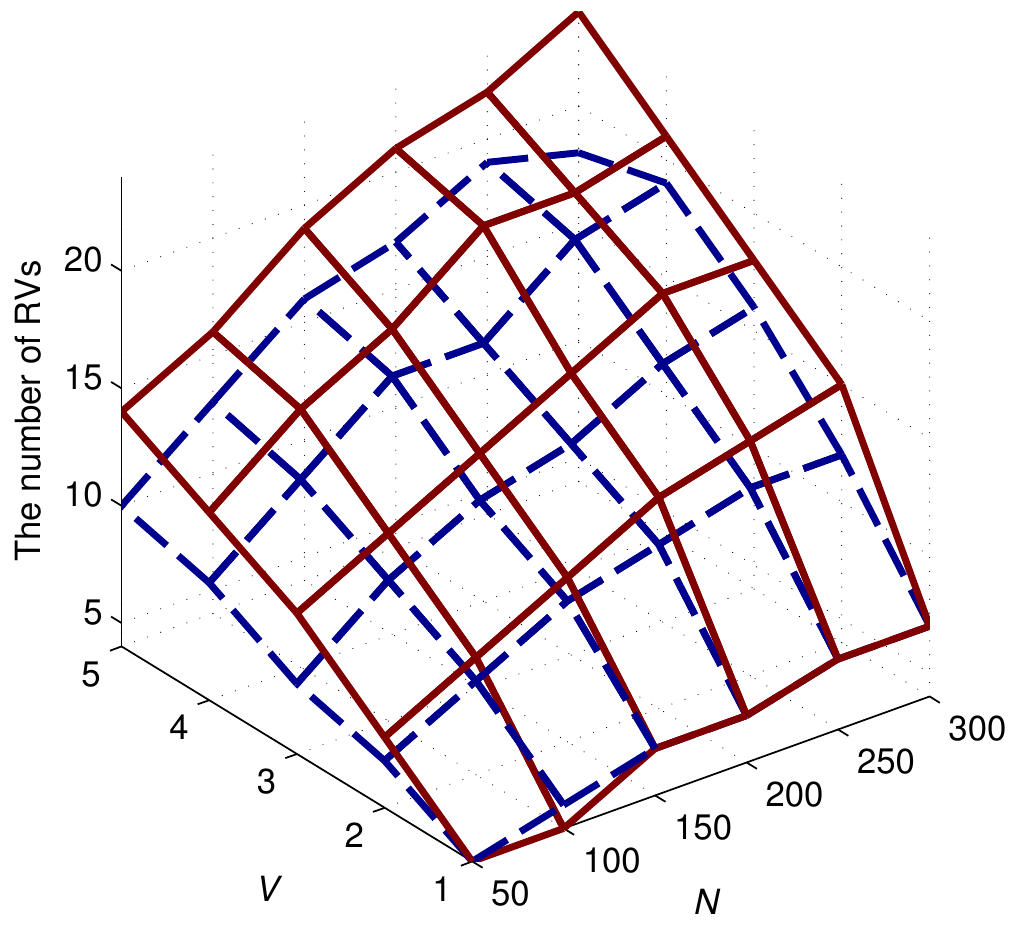}
            \caption{The number of RVs}
            \label{fig:mesh:f}
    \end{subfigure}
    \caption{Median values of MC simulations (the number of simulations is 101)}
    \label{fig:mesh}
\end{figure}

Fig.~\ref{fig:mesh} shows the median values of the performance measures of both methods with various $V$ and $N$ values. Entropy loss, quadratic loss, and RMSE decrease as $N$ increases: the greater the number of training samples, the more accurate the estimation. In contrast, the number of RVs, the number of iterations of each EM algorithm (the same tolerance value of 0.1 for checking the convergence of each EM algorithm was used as in 43th line of Algorithm~\ref{alg:exi} and 30th line of Algorithm~\ref{alg:pro}), and the running time (only for learning without prediction) of each EM algorithm  increase as $N$ increases: the greater the number of training samples, the greater the computational burden.

\input{table}

Tables~\ref{tab:time}\textendash\ref{tab:number} show \rmnum1) the difference in the median values of the performance measures, where each median value is obtained from 101 MC simulations, and then the median value of the proposed method is subtracted from that of the existing method (i.e. positive difference values mean that the proposed method is better than the existing method, while negative difference values mean the opposite) and \rmnum2) the \textit{p}-values of the Wilcoxon rank sum tests, which appear inside the brackets. The \mbox{\textit{p}-value} is interpreted as the probability that a difference in the median values would be obtained given that the population medians of two methods are equivalent, i.e. the \textit{p}-value is not equal to the probability that the population medians are equivalent~\cite[p.~635]{simon2013evolutionary}. Note that statistically significant difference values are marked in bold (\textit{p}-value $<0.05$).

The proposed method is faster than the existing method as shown in Table~\ref{tab:time} (all differences are statistically significant). In particular, the time difference is amplified as $V$ or $N$ increases. This is because the time complexity of the proposed method $O\left(V^3+M^3\right)$ is less than that of the existing method $O\left(VM^3\right)$ (\mbox{$O\left(V^3+M^3\right)<O\left(VM^3\right)$} is satisfied since $V<M$ is satisfied in most applications). Note that even when the number of input dimensions $U$ changes, the size of the design matrix $\mathbf\Phi$ does not change. Hence, $U$ does not influence the time complexity of both the methods.

Furthermore, the proposed method achieves higher accuracy in estimating the covariance matrix of the noise $\mathbf\Omega$ than the existing method as shown in Table~\ref{tab:entropy} and Table~\ref{tab:quad}. This is because the proposed method considers the correlation matrix of the noise as Eq.~\eqref{eq:pro:p(T|W,O)}, but the existing method does not as Eq.~\eqref{eq:pro:p(T|W,O)}.

However, the proposed method is worse than the existing one in terms of \rmnum1) the accuracy in predicting the mean values as shown in Table~\ref{tab:rmse} (in particular, the RMSE increases in the region of high $V$ and low $N$) and \rmnum2) the number of RVs as shown in Table~\ref{tab:number}. This is because the proposed method has the assumption of the weight $\mathbf\Omega=\frac{\mathbb E\left[\mathbf W^\intercal \mathbf W\right]}{\textnormal{tr}\left(\mathbf A^{-1}\right)}$, which behaves as the constraint of the weight. Consequently, the mean values tend to deviate from the true functions, and the number of RVs increases.

The MATLAB codes of the experiment have been uploaded on \url{http://www.mathworks.com/matlabcentral/fileexchange/49131} to avoid any potential ambiguity of both the methods. %\footnote{For the matrix inversion operation, Cholesky factorisation is used to compute the covariance matrix $\mathbf\Sigma_j$ in Eq.~(\ref{eq:exi:Sigma})~\cite{tipping2009efficient}, but MATLAB operator \textbackslash, which solves systems of linear equations, is used to compute $\mathbf\Sigma$ in Eq.~(\ref{eq:pro:Sigma}).}

\section{Conclusion}\label{sec:conclusion}
A new algorithm of MRVR has been proposed. It is more efficient in computing the weight $\mathbf W$ and more accurate in estimating the covariance matrix of the noise $\mathbf\Omega$ than the existing algorithm. Its computational efficiency and accuracy can be attributed to the different model specifications: the existing method expresses the likelihood of the training data as the product of the Gaussian distributions in Eq.~(\ref{eq:exi:p(T|W,s)}), but the proposed one expresses it as the matrix Gaussian distribution in Eq.~(\ref{eq:pro:p(T|W,O)}).

However, the proposed method has drawbacks of lower accuracy in estimating the mean of the weight $\mathbf M$ in Eq.~(\ref{eq:pro:M}) and higher number of RVs than the existing method. These disadvantages are caused by the assumption $\mathbf\Omega=\frac{\mathbb E\left[\mathbf W^\intercal \mathbf W\right]}{\textnormal{tr}\left(\mathbf A^{-1}\right)}$, which means the weight $\mathbf W$ is related to the noise $\mathbf E$ in Eq.~(\ref{eq:pro:T}), but it was indispensable to make MRVR faster.

\bibliographystyle{rQUF}
\bibliography{template}

%\begin{figure}
%% Use the relevant command to insert your figure file.
%% For example, with the graphicx package use
%  \includegraphics{example.eps}
%\label{fig:ex}       % Give a unique label
%\end{figure}
%%
%% For two-column wide figures use
%\begin{figure*}
%% Use the relevant command to insert your figure file.
%% For example, with the graphicx package use
%  \includegraphics[width=0.75\textwidth]{example.eps}
%\end{figure*}
%\begin{table}
%\begin{tabular}{lll}
%\hline\noalign{\smallskip}
%\end{tabular}
%\end{table}

%\begin{acknowledgements}
%The author would like to thank
%\end{acknowledgements}
%\end{linenumbers}
\appendix
\setcounter{secnumdepth}{0}
\section{Appendix: proof of Eq.~(\ref{eq:pro:p(W|T,a,O)}) and Eq.~(\ref{eq:pro:p(T|a,O)})}
\begin{equation*}
\begin{split}
&p\left(\mathbf W|\mathbf T,\bm\upalpha,\mathbf\Omega \right)p\left(\mathbf T|\bm\upalpha,\mathbf\Omega\right)\\
=&p\left(\mathbf T|\mathbf W,\mathbf\Omega \right)p\left(\mathbf W|\bm\upalpha,\mathbf\Omega\right)\\
=&\rho\textnormal{exp}\left(-\frac{1}{2}\textnormal{tr}\left(\mathbf\Omega ^{-1}\left(\mathbf T-\mathbf\Phi \mathbf W\right)^\intercal\left(\mathbf T-\mathbf\Phi \mathbf W\right)\right)\right)\textnormal{exp}\left(-\frac{1} 2\textnormal{tr}\left(\mathbf\Omega
^{-1}\mathbf W^\intercal \mathbf A\mathbf W\right)\right),\\
&\textnormal{where }\rho=\left(2\pi \right)^{-\frac{VN} 2}\left|\mathbf\Omega
\right|^{-\frac{N} 2}\left(2\pi \right)^{-\frac{V\left(N+1\right)}
2}\left|\mathbf\Omega \right|^{-\frac{N+1} 2}\left|\mathbf A \right|^{\frac{V} 2}\\
=&\rho\textnormal{exp}\left(-\frac{1} 2\textnormal{tr}\left(\mathbf\Omega ^{-1}\left(\left(\mathbf T^\intercal-\mathbf W^\intercal\mathbf\Phi^\intercal\right)\left(\mathbf T-\mathbf\Phi
\mathbf W\right)+\mathbf W^\intercal \mathbf A \mathbf W\right)\right)\right)\\
=&\rho\textnormal{exp}\left(-\frac{1} 2\textnormal{tr}\left(\mathbf\Omega ^{-1}\left(\mathbf T^\intercal \mathbf T-\mathbf T^\intercal\mathbf\Phi \mathbf W-\mathbf W^\intercal\mathbf\Phi ^\intercal\mathbf T+\mathbf W^\intercal\mathbf\Sigma^{-1}\mathbf W\right)\right)\right),\textnormal{ where }\mathbf\Sigma=\left(\mathbf\Phi ^\intercal\mathbf\Phi
+\mathbf A\right)^{-1}\\
=&\rho\textnormal{exp}\left(-\frac{1} 2\textnormal{tr}\left(\mathbf\Omega ^{-1}\left(\mathbf T^\intercal\mathbf T-\mathbf T^\intercal\mathbf\Phi \mathbf W+\mathbf W^\intercal\mathbf\Sigma
^{-1}\left(\mathbf W-\mathbf M\right)\right)\right)\right),\textnormal{where }\mathbf M=\mathbf\Sigma \mathbf\Phi^\intercal\mathbf T\\
=&\rho\textnormal{exp}\left(-\frac{1} 2\textnormal{tr}\left(\mathbf\Omega ^{-1}\left(\mathbf T^\intercal\mathbf T-\mathbf T^\intercal\mathbf\Phi \mathbf W+\mathbf W^\intercal\mathbf\Sigma
^{-1}\left(\mathbf W-\mathbf M\right)- \mathbf M^\intercal\mathbf\Sigma
^{-1}\left(\mathbf W-\mathbf M\right)+ \mathbf M^\intercal\mathbf\Sigma
^{-1}\left(\mathbf W-\mathbf M\right)\right)\right)\right)\\
=&\rho\textnormal{exp}\left(-\frac{1} 2\textnormal{tr}\left(\mathbf\Omega ^{-1}\left(\left(\mathbf W-\mathbf M\right)^\intercal\mathbf\Sigma ^{-1}\left(\mathbf W-\mathbf M\right)+\mathbf M^\intercal\mathbf\Sigma
^{-1}\left(\mathbf W-\mathbf M\right)+\mathbf T^\intercal\mathbf T-\mathbf T^\intercal\mathbf\Phi \mathbf W\right)\right)\right)\\
=&\rho\textnormal{exp}\left(-\frac{1} 2\textnormal{tr}\left(\mathbf\Omega ^{-1}\left(\left(\mathbf W-\mathbf M\right)^\intercal\mathbf\Sigma ^{-1}\left(\mathbf W-\mathbf M\right)+\mathbf T^\intercal\mathbf\Phi
\left(\mathbf W-\mathbf M\right)+\mathbf T^\intercal \mathbf T-\mathbf T^\intercal\mathbf\Phi \mathbf W\right)\right)\right),\\
&\textnormal{since }\mathbf\Sigma=\mathbf\Sigma^\intercal\\
=&\rho\textnormal{exp}\left(-\frac{1} 2\textnormal{tr}\left(\mathbf\Omega ^{-1}\left(\left(\mathbf W-\mathbf M\right)^\intercal\mathbf\Sigma ^{-1}\left(\mathbf W-\mathbf M\right)+\mathbf T^\intercal\left(\mathbf T-\mathbf\Phi
\mathbf M\right)\right)\right)\right)\\
=&\rho\textnormal{exp}\left(-\frac{1} 2\textnormal{tr}\left(\mathbf\Omega ^{-1}\left(\left(\mathbf W-\mathbf M\right)^\intercal\mathbf\Sigma ^{-1}\left(\mathbf W-\mathbf M\right)+\mathbf T^\intercal\left(\mathbf I-\mathbf\Phi
\left(\mathbf\Phi ^\intercal\mathbf\Phi +\mathbf A\right)^{-1}\mathbf\Phi ^\intercal\right)\mathbf T\right)\right)\right)\\
=&\rho\textnormal{exp}\left(-\frac{1} 2\textnormal{tr}\left(\mathbf\Omega ^{-1}\left(\left(\mathbf W-\mathbf M\right)^\intercal\mathbf\Sigma ^{-1}\left(\mathbf W-\mathbf M\right)+\mathbf T^\intercal\left(\mathbf I+\mathbf\Phi
\mathbf A^{-1}\mathbf\Phi ^\intercal\right)^{-1}\mathbf T\right)\right)\right),\\
&\textnormal{by the Woodbury matrix identity}\\
=&\left(2\pi \right)^{-\frac{V\left(N+1\right)} 2}\left|\mathbf\Omega \right|^{-\frac{N+1} 2}\left|\mathbf\Sigma \right|^{-\frac{V}
2}\textnormal{exp}\left(-\frac{1} 2\textnormal{tr}\left(\mathbf\Omega ^{-1}\left(\mathbf W-\mathbf M\right)^\intercal\mathbf\Sigma ^{-1}\left(\mathbf W-\mathbf M\right)\right)\right)\\
&\left(2\pi
\right)^{-\frac{VN} 2}\left|\mathbf\Omega \right|^{-\frac{N} 2}\left|\mathbf I+\mathbf\Phi\mathbf A^{-1}\mathbf\Phi ^\intercal\right|^{-\frac{V}
2}\textnormal{exp}\left(-\frac{1} 2\textnormal{tr}\left(\mathbf\Omega ^{-1}\mathbf T^\intercal\left(\mathbf I+\mathbf\Phi \mathbf A^{-1}\mathbf\Phi^\intercal\right)^{-1}\mathbf T\right)\right)
\end{split}
\end{equation*}
\end{document}

%% file: table.tex
\begin{table}\caption{The difference in median values of running time (seconds)}\centering\begin{tabular}{ccccccc}\hline  & $N=50$ & $N=100$ & $N=150$ & $N=200$ & $N=250$ & $N=300$ \\\hline $V=1$ & $\makecell{\mathbf{0.04}\\\left(3.6\times10^{\mbox{-}10}\right)}$ & $\makecell{\mathbf{0.19}\\\left(2.1\times10^{\mbox{-}9}\right)}$ & $\makecell{\mathbf{0.40}\\\left(4.9\times10^{\mbox{-}9}\right)}$ & $\makecell{\mathbf{1.02}\\\left(3.2\times10^{\mbox{-}9}\right)}$ & $\makecell{\mathbf{3.11}\\\left(1.8\times10^{\mbox{-}20}\right)}$ & $\makecell{\mathbf{5.95}\\\left(7.1\times10^{\mbox{-}18}\right)}$ \\$V=2$ & $\makecell{\mathbf{0.20}\\\left(4.8\times10^{\mbox{-}25}\right)}$ & $\makecell{\mathbf{0.76}\\\left(1.5\times10^{\mbox{-}22}\right)}$ & $\makecell{\mathbf{2.44}\\\left(9.0\times10^{\mbox{-}26}\right)}$ & $\makecell{\mathbf{5.95}\\\left(2.1\times10^{\mbox{-}25}\right)}$ & $\makecell{\mathbf{17.20}\\\left(1.3\times10^{\mbox{-}30}\right)}$ & $\makecell{\mathbf{26.17}\\\left(1.0\times10^{\mbox{-}31}\right)}$ \\$V=3$ & $\makecell{\mathbf{0.33}\\\left(3.2\times10^{\mbox{-}26}\right)}$ & $\makecell{\mathbf{1.70}\\\left(3.8\times10^{\mbox{-}32}\right)}$ & $\makecell{\mathbf{5.27}\\\left(5.4\times10^{\mbox{-}29}\right)}$ & $\makecell{\mathbf{10.37}\\\left(1.1\times10^{\mbox{-}31}\right)}$ & $\makecell{\mathbf{37.74}\\\left(4.7\times10^{\mbox{-}34}\right)}$ & $\makecell{\mathbf{64.99}\\\left(2.2\times10^{\mbox{-}34}\right)}$ \\$V=4$ & $\makecell{\mathbf{0.74}\\\left(2.2\times10^{\mbox{-}33}\right)}$ & $\makecell{\mathbf{3.60}\\\left(1.3\times10^{\mbox{-}32}\right)}$ & $\makecell{\mathbf{9.69}\\\left(2.1\times10^{\mbox{-}33}\right)}$ & $\makecell{\mathbf{24.51}\\\left(2.4\times10^{\mbox{-}33}\right)}$ & $\makecell{\mathbf{57.14}\\\left(2.1\times10^{\mbox{-}34}\right)}$ & $\makecell{\mathbf{112.18}\\\left(1.2\times10^{\mbox{-}34}\right)}$ \\$V=5$ & $\makecell{\mathbf{1.05}\\\left(1.1\times10^{\mbox{-}33}\right)}$ & $\makecell{\mathbf{5.25}\\\left(5.4\times10^{\mbox{-}34}\right)}$ & $\makecell{\mathbf{13.01}\\\left(6.3\times10^{\mbox{-}34}\right)}$ & $\makecell{\mathbf{34.09}\\\left(2.2\times10^{\mbox{-}34}\right)}$ & $\makecell{\mathbf{92.96}\\\left(1.2\times10^{\mbox{-}34}\right)}$ & $\makecell{\mathbf{176.24}\\\left(1.2\times10^{\mbox{-}34}\right)}$ \\\hline \end{tabular}\label{tab:time}\end{table}\begin{table}\caption{The difference in median values of entropy loss}\centering\begin{tabular}{ccccccc}\hline  & $N=50$ & $N=100$ & $N=150$ & $N=200$ & $N=250$ & $N=300$ \\\hline $V=1$ & $\makecell{3.4\times10^{\mbox{-}4}\\\left(0.9904\right)}$ & $\makecell{8.0\times10^{\mbox{-}5}\\\left(0.9176\right)}$ & $\makecell{2.2\times10^{\mbox{-}5}\\\left(0.9962\right)}$ & $\makecell{1.7\times10^{\mbox{-}5}\\\left(0.9520\right)}$ & $\makecell{3.7\times10^{\mbox{-}5}\\\left(0.8454\right)}$ & $\makecell{5.7\times10^{\mbox{-}5}\\\left(0.9808\right)}$ \\$V=2$ & $\makecell{3.9\times10^{\mbox{-}3}\\\left(0.4942\right)}$ & $\makecell{3.3\times10^{\mbox{-}3}\\\left(0.5964\right)}$ & $\makecell{3.4\times10^{\mbox{-}3}\\\left(0.1132\right)}$ & $\makecell{6.8\times10^{\mbox{-}4}\\\left(0.3915\right)}$ & $\makecell{1.1\times10^{\mbox{-}3}\\\left(0.7434\right)}$ & $\makecell{4.0\times10^{\mbox{-}4}\\\left(0.6234\right)}$ \\$V=3$ & $\makecell{3.2\times10^{\mbox{-}2}\\\left(0.0883\right)}$ & $\makecell{8.6\times10^{\mbox{-}3}\\\left(0.2469\right)}$ & $\makecell{\mathbf{6.8\times10^{\mbox{-}3}}\\\left(0.0192\right)}$ & $\makecell{\mathbf{8.0\times10^{\mbox{-}3}}\\\left(0.0069\right)}$ & $\makecell{\mathbf{5.1\times10^{\mbox{-}3}}\\\left(0.0119\right)}$ & $\makecell{\mathbf{1.4\times10^{\mbox{-}3}}\\\left(0.0412\right)}$ \\$V=4$ & $\makecell{3.9\times10^{\mbox{-}2}\\\left(0.1137\right)}$ & $\makecell{\mathbf{3.0\times10^{\mbox{-}2}}\\\left(0.0030\right)}$ & $\makecell{\mathbf{1.5\times10^{\mbox{-}2}}\\\left(0.0010\right)}$ & $\makecell{\mathbf{1.1\times10^{\mbox{-}2}}\\\left(0.0042\right)}$ & $\makecell{\mathbf{9.3\times10^{\mbox{-}3}}\\\left(0.0063\right)}$ & $\makecell{\mathbf{4.4\times10^{\mbox{-}3}}\\\left(0.0058\right)}$ \\$V=5$ & $\makecell{\mathbf{1.1\times10^{\mbox{-}1}}\\\left(0.0231\right)}$ & $\makecell{\mathbf{3.0\times10^{\mbox{-}2}}\\\left(0.0015\right)}$ & $\makecell{\mathbf{2.9\times10^{\mbox{-}2}}\\\left(0.0000\right)}$ & $\makecell{\mathbf{2.5\times10^{\mbox{-}2}}\\\left(0.0008\right)}$ & $\makecell{\mathbf{1.4\times10^{\mbox{-}2}}\\\left(0.0005\right)}$ & $\makecell{\mathbf{1.2\times10^{\mbox{-}2}}\\\left(0.0018\right)}$ \\\hline \end{tabular}\label{tab:entropy}\end{table}\begin{table}\caption{The difference in median values of quadratic loss}\centering\begin{tabular}{ccccccc}\hline  & $N=50$ & $N=100$ & $N=150$ & $N=200$ & $N=250$ & $N=300$ \\\hline $V=1$ & $\makecell{2.8\times10^{\mbox{-}3}\\\left(0.9981\right)}$ & $\makecell{\mbox{-}5.7\times10^{\mbox{-}4}\\\left(0.9233\right)}$ & $\makecell{4.9\times10^{\mbox{-}5}\\\left(0.9981\right)}$ & $\makecell{2.3\times10^{\mbox{-}4}\\\left(0.9405\right)}$ & $\makecell{4.4\times10^{\mbox{-}5}\\\left(0.8605\right)}$ & $\makecell{2.0\times10^{\mbox{-}4}\\\left(0.9770\right)}$ \\$V=2$ & $\makecell{5.5\times10^{\mbox{-}3}\\\left(0.5080\right)}$ & $\makecell{3.5\times10^{\mbox{-}3}\\\left(0.4526\right)}$ & $\makecell{7.0\times10^{\mbox{-}3}\\\left(0.0915\right)}$ & $\makecell{2.2\times10^{\mbox{-}3}\\\left(0.3616\right)}$ & $\makecell{1.6\times10^{\mbox{-}3}\\\left(0.6824\right)}$ & $\makecell{\mbox{-}1.9\times10^{\mbox{-}4}\\\left(0.5782\right)}$ \\$V=3$ & $\makecell{\mathbf{8.5\times10^{\mbox{-}2}}\\\left(0.0180\right)}$ & $\makecell{2.5\times10^{\mbox{-}2}\\\left(0.1154\right)}$ & $\makecell{\mathbf{1.4\times10^{\mbox{-}2}}\\\left(0.0109\right)}$ & $\makecell{\mathbf{1.4\times10^{\mbox{-}2}}\\\left(0.0029\right)}$ & $\makecell{\mathbf{1.3\times10^{\mbox{-}2}}\\\left(0.0051\right)}$ & $\makecell{\mathbf{2.7\times10^{\mbox{-}3}}\\\left(0.0263\right)}$ \\$V=4$ & $\makecell{\mathbf{1.7\times10^{\mbox{-}1}}\\\left(0.0382\right)}$ & $\makecell{\mathbf{8.6\times10^{\mbox{-}2}}\\\left(0.0005\right)}$ & $\makecell{\mathbf{5.4\times10^{\mbox{-}2}}\\\left(0.0001\right)}$ & $\makecell{\mathbf{2.4\times10^{\mbox{-}2}}\\\left(0.0007\right)}$ & $\makecell{\mathbf{2.9\times10^{\mbox{-}2}}\\\left(0.0019\right)}$ & $\makecell{\mathbf{1.5\times10^{\mbox{-}2}}\\\left(0.0013\right)}$ \\$V=5$ & $\makecell{\mathbf{7.3\times10^{\mbox{-}1}}\\\left(0.0058\right)}$ & $\makecell{\mathbf{1.8\times10^{\mbox{-}1}}\\\left(0.0000\right)}$ & $\makecell{\mathbf{8.5\times10^{\mbox{-}2}}\\\left(0.0000\right)}$ & $\makecell{\mathbf{5.5\times10^{\mbox{-}2}}\\\left(0.0001\right)}$ & $\makecell{\mathbf{4.8\times10^{\mbox{-}2}}\\\left(0.0000\right)}$ & $\makecell{\mathbf{3.7\times10^{\mbox{-}2}}\\\left(0.0004\right)}$ \\\hline \end{tabular}\label{tab:quad}\end{table}\begin{table}\caption{The difference in median values of RMSE}\centering\begin{tabular}{ccccccc}\hline  & $N=50$ & $N=100$ & $N=150$ & $N=200$ & $N=250$ & $N=300$ \\\hline $V=1$ & $\makecell{\mbox{-}0.0038\\\left(0.956\right)}$ & $\makecell{\mbox{-}0.0000\\\left(0.985\right)}$ & $\makecell{0.0000\\\left(0.967\right)}$ & $\makecell{\mbox{-}0.0001\\\left(1.000\right)}$ & $\makecell{\mbox{-}0.0007\\\left(0.965\right)}$ & $\makecell{0.0001\\\left(0.987\right)}$ \\$V=2$ & $\makecell{\mbox{-}0.0035\\\left(0.898\right)}$ & $\makecell{\mbox{-}0.0076\\\left(0.544\right)}$ & $\makecell{\mbox{-}0.0051\\\left(0.562\right)}$ & $\makecell{\mbox{-}0.0033\\\left(0.758\right)}$ & $\makecell{\mbox{-}0.0064\\\left(0.646\right)}$ & $\makecell{\mbox{-}0.0020\\\left(0.977\right)}$ \\$V=3$ & $\makecell{\mbox{-}0.0177\\\left(0.182\right)}$ & $\makecell{\mbox{-}0.0049\\\left(0.299\right)}$ & $\makecell{\mbox{-}0.0012\\\left(0.408\right)}$ & $\makecell{\mbox{-}0.0030\\\left(0.661\right)}$ & $\makecell{\mbox{-}0.0013\\\left(0.546\right)}$ & $\makecell{\mbox{-}0.0037\\\left(0.565\right)}$ \\$V=4$ & $\makecell{\mathbf{\mbox{-}0.0121}\\\left(0.041\right)}$ & $\makecell{\mathbf{\mbox{-}0.0142}\\\left(0.034\right)}$ & $\makecell{\mathbf{\mbox{-}0.0101}\\\left(0.017\right)}$ & $\makecell{\mbox{-}0.0037\\\left(0.092\right)}$ & $\makecell{\mbox{-}0.0038\\\left(0.115\right)}$ & $\makecell{\mbox{-}0.0029\\\left(0.293\right)}$ \\$V=5$ & $\makecell{\mathbf{\mbox{-}0.0150}\\\left(0.008\right)}$ & $\makecell{\mathbf{\mbox{-}0.0026}\\\left(0.013\right)}$ & $\makecell{\mathbf{\mbox{-}0.0149}\\\left(0.001\right)}$ & $\makecell{\mathbf{\mbox{-}0.0110}\\\left(0.015\right)}$ & $\makecell{\mathbf{\mbox{-}0.0068}\\\left(0.025\right)}$ & $\makecell{\mbox{-}0.0062\\\left(0.072\right)}$ \\\hline \end{tabular}\label{tab:rmse}\end{table}\begin{table}\caption{The difference in median values of the number of RVs}\centering\begin{tabular}{ccccccc}\hline  & $N=50$ & $N=100$ & $N=150$ & $N=200$ & $N=250$ & $N=300$ \\\hline $V=1$ & $\makecell{0\\\left(8.8\times10^{\mbox{-}1}\right)}$ & $\makecell{1\\\left(9.0\times10^{\mbox{-}1}\right)}$ & $\makecell{0\\\left(8.4\times10^{\mbox{-}1}\right)}$ & $\makecell{0\\\left(9.6\times10^{\mbox{-}1}\right)}$ & $\makecell{0\\\left(8.9\times10^{\mbox{-}1}\right)}$ & $\makecell{0\\\left(9.9\times10^{\mbox{-}1}\right)}$ \\$V=2$ & $\makecell{\mbox{-}1\\\left(5.2\times10^{\mbox{-}2}\right)}$ & $\makecell{\mathbf{\mbox{-}1}\\\left(4.8\times10^{\mbox{-}4}\right)}$ & $\makecell{\mathbf{\mbox{-}1}\\\left(3.2\times10^{\mbox{-}3}\right)}$ & $\makecell{\mathbf{\mbox{-}2}\\\left(5.1\times10^{\mbox{-}4}\right)}$ & $\makecell{\mathbf{\mbox{-}2}\\\left(1.5\times10^{\mbox{-}3}\right)}$ & $\makecell{\mathbf{\mbox{-}3}\\\left(2.2\times10^{\mbox{-}3}\right)}$ \\$V=3$ & $\makecell{\mathbf{\mbox{-}3}\\\left(3.6\times10^{\mbox{-}7}\right)}$ & $\makecell{\mathbf{\mbox{-}2}\\\left(3.9\times10^{\mbox{-}5}\right)}$ & $\makecell{\mathbf{\mbox{-}2}\\\left(1.5\times10^{\mbox{-}5}\right)}$ & $\makecell{\mathbf{\mbox{-}3}\\\left(3.4\times10^{\mbox{-}7}\right)}$ & $\makecell{\mathbf{\mbox{-}3}\\\left(3.2\times10^{\mbox{-}6}\right)}$ & $\makecell{\mathbf{\mbox{-}2}\\\left(2.6\times10^{\mbox{-}6}\right)}$ \\$V=4$ & $\makecell{\mathbf{\mbox{-}3}\\\left(2.2\times10^{\mbox{-}8}\right)}$ & $\makecell{\mathbf{\mbox{-}3}\\\left(4.8\times10^{\mbox{-}6}\right)}$ & $\makecell{\mathbf{\mbox{-}2}\\\left(8.6\times10^{\mbox{-}5}\right)}$ & $\makecell{\mathbf{\mbox{-}5}\\\left(2.4\times10^{\mbox{-}9}\right)}$ & $\makecell{\mathbf{\mbox{-}2}\\\left(2.2\times10^{\mbox{-}7}\right)}$ & $\makecell{\mathbf{\mbox{-}2}\\\left(1.8\times10^{\mbox{-}4}\right)}$ \\$V=5$ & $\makecell{\mathbf{\mbox{-}4}\\\left(3.9\times10^{\mbox{-}15}\right)}$ & $\makecell{\mathbf{\mbox{-}3}\\\left(3.9\times10^{\mbox{-}10}\right)}$ & $\makecell{\mathbf{\mbox{-}3}\\\left(3.9\times10^{\mbox{-}8}\right)}$ & $\makecell{\mathbf{\mbox{-}4}\\\left(9.0\times10^{\mbox{-}10}\right)}$ & $\makecell{\mathbf{\mbox{-}3}\\\left(6.9\times10^{\mbox{-}7}\right)}$ & $\makecell{\mathbf{\mbox{-}6}\\\left(3.2\times10^{\mbox{-}7}\right)}$ \\\hline \end{tabular}\label{tab:number}\end{table}

%% file: FMRVR.bbl
\begin{thebibliography}{29}
\providecommand{\natexlab}[1]{#1}
\providecommand{\noopsort}[1]{}
\providecommand{\printfirst}[2]{#1}
\providecommand{\singleletter}[1]{#1}
\providecommand{\switchargs}[2]{#2#1}

\bibitem[\protect\citeauthoryear{Alvarez and
  Lawrence}{2009}]{alvarez2009sparse}
Alvarez, M. and Lawrence, N.D., Sparse convolved Gaussian processes for
  multi-output regression. In {\itshape Proceedings of the }{\itshape Advances
  in Neural Information Processing Systems}, pp. 57--64, 2009.

\bibitem[\protect\citeauthoryear{Anderson}{1984}]{anderson1984introduction}
Anderson, T.W., {\itshape An introduction to multivariate statistical analysis}
  (2  edn), 1984, Wiley.

\bibitem[\protect\citeauthoryear{Arnold}{1981}]{arnold1981theory}
Arnold, S.F., {\itshape The theory of linear models and multivariate analysis},
  1981, Wiley.

\bibitem[\protect\citeauthoryear{Ben-Shimon and
  Shmilovici}{2006}]{ben2006accelerating}
Ben-Shimon, D. and Shmilovici, A., Accelerating the relevance vector machine
  via data partitioning. {\itshape Foundations of Computing and Decision
  Sciences}, 2006, \textbf{31}, 27--41.

\bibitem[\protect\citeauthoryear{Bonilla
  {\itshape{et~al.}}}{2007}]{bonilla2007multi}
Bonilla, E.V., Chai, K.M.A. and Williams, C.K.I., Multi-task Gaussian process
  prediction. In {\itshape Proceedings of the }{\itshape Advances in Neural
  Information Processing Systems}, pp. 153--160, 2007.

\bibitem[\protect\citeauthoryear{Boyle and Frean}{2004}]{boyle2004dependent}
Boyle, P. and Frean, M.R., Dependent Gaussian Processes.. In {\itshape
  Proceedings of the }{\itshape Advances in Neural Information Processing
  Systems}, pp. 217--224, 2004.

\bibitem[\protect\citeauthoryear{Catanzaro
  {\itshape{et~al.}}}{2008}]{catanzaro2008fast}
Catanzaro, B., Sundaram, N. and Keutzer, K., Fast support vector machine
  training and classification on graphics processors. In {\itshape Proceedings
  of the }{\itshape Proceedings of the 25th international conference on Machine
  learning}, pp. 104--111, 2008.

\bibitem[\protect\citeauthoryear{Chang and Lin}{2011}]{chang2011libsvm}
Chang, C.C. and Lin, C.J., LIBSVM: A library for support vector machines.
  {\itshape ACM Transactions on Intelligent Systems and Technology}, 2011,
  \textbf{2}, 27:1--27.

\bibitem[\protect\citeauthoryear{Chu
  {\itshape{et~al.}}}{2004}]{chu2004bayesian}
Chu, W., Keerthi, S.S. and Ong, C.J., Bayesian support vector regression using
  a unified loss function. {\itshape IEEE Transactions on Neural Networks},
  2004, \textbf{15}, 29--44.

\bibitem[\protect\citeauthoryear{Cortes and Vapnik}{1995}]{cortes1995support}
Cortes, C. and Vapnik, V., Support-vector networks. {\itshape Machine
  learning}, 1995, \textbf{20}, 273--297.

\bibitem[\protect\citeauthoryear{Gao
  {\itshape{et~al.}}}{2002}]{gao2002probabilistic}
Gao, J.B., Gunn, S.R., Harris, C.J. and Brown, M., A probabilistic framework
  for SVM regression and error bar estimation. {\itshape Machine Learning},
  2002, \textbf{46}, 71--89.

\bibitem[\protect\citeauthoryear{Gibbs}{1997}]{gibbs1997bayesian}
Gibbs, M., Bayesian Gaussian processes for classification and regression. PhD
  thesis, University of Cambridge, 1997.

\bibitem[\protect\citeauthoryear{Gramacy
  {\itshape{et~al.}}}{2014}]{gramacy2014massively}
Gramacy, R.B., Niemi, J. and Weiss, R.M., Massively parallel approximate
  Gaussian process regression. {\itshape SIAM/ASA Journal on Uncertainty
  Quantification}, 2014, \textbf{2}, 564--584.

\bibitem[\protect\citeauthoryear{Guo and Zhang}{2007}]{guo2007reducing}
Guo, G. and Zhang, J.S., Reducing examples to accelerate support vector
  regression. {\itshape Pattern Recognition Letters}, 2007, \textbf{28},
  2173--2183.

\bibitem[\protect\citeauthoryear{Montgomery}{2013}]{montgomery2013applied}
Montgomery, D.C., {\itshape Applied Statistics and Probability for Engineers}
  (6  edn), 2013, Wiley.

\bibitem[\protect\citeauthoryear{P{\'e}rez-Cruz
  {\itshape{et~al.}}}{2002}]{perez2002multi}
P{\'e}rez-Cruz, F., Camps-Valls, G., Soria-Olivas, E., P{\'e}rez-Ruixo, J.J.,
  Figueiras-Vidal, A.R. and Art{\'e}s-Rodr{\'\i}guez, A., Multi-dimensional
  function approximation and regression estimation. In {\itshape Proceedings of
  the }{\itshape International Conference on Artificial Neural Networks}, pp.
  757--762, 2002.

\bibitem[\protect\citeauthoryear{Sch{\"o}lkopf
  {\itshape{et~al.}}}{2000}]{scholkopf2000new}
Sch{\"o}lkopf, B., Smola, A.J., Williamson, R.C. and Bartlett, P.L., New
  support vector algorithms. {\itshape Neural computation}, 2000, \textbf{12},
  1207--1245.

\bibitem[\protect\citeauthoryear{Seeger
  {\itshape{et~al.}}}{2003}]{seeger2003fast}
Seeger, M., Williams, C. and Lawrence, N., Fast forward selection to speed up
  sparse Gaussian process regression. In {\itshape Proceedings of the
  }{\itshape Workshop on Artificial Intelligence and Statistics 9}, 2003.

\bibitem[\protect\citeauthoryear{Shen {\itshape{et~al.}}}{2006}]{shen2006fast}
Shen, Y., Ng, A. and Seeger, M., Fast Gaussian process regression using
  kd-trees. In {\itshape Proceedings of the }{\itshape 19th Annual Conference
  on Neural Information Processing Systems}, 2006.

\bibitem[\protect\citeauthoryear{Simon}{2013}]{simon2013evolutionary}
Simon, D., {\itshape Evolutionary optimization algorithms}, 2013, Wiley.

\bibitem[\protect\citeauthoryear{Srinivasan
  {\itshape{et~al.}}}{2010}]{srinivasan2010gpuml}
Srinivasan, B.V., Qi, H. and Duraiswami, R., GPUML: Graphical processors for
  speeding up kernel machines. In {\itshape Proceedings of the }{\itshape
  Workshop on High Performance Analytics - Algorithms, Implementations, and
  Applications}, 2010.

\bibitem[\protect\citeauthoryear{Thayananthan}{2005}]{thayananthan2005template}
Thayananthan, A., Template-based pose estimation and tracking of 3D hand
  motion. PhD thesis, University of Cambridge, 2005.

\bibitem[\protect\citeauthoryear{Thayananthan
  {\itshape{et~al.}}}{2008}]{thayananthan2008pose}
Thayananthan, A., Navaratnam, R., Stenger, B., Torr, P.H. and Cipolla, R., Pose
  estimation and tracking using multivariate regression. {\itshape Pattern
  Recognition Letters}, 2008, \textbf{29}, 1302--1310.

\bibitem[\protect\citeauthoryear{Tipping}{2001}]{tipping2001sparse}
Tipping, M.E., Sparse Bayesian learning and the relevance vector machine.
  {\itshape The journal of machine learning research}, 2001, \textbf{1},
  211--244.

\bibitem[\protect\citeauthoryear{Tipping and Faul}{2003}]{tipping2003fast}
Tipping, M.E. and Faul, A.C., Fast marginal likelihood maximisation for sparse
  Bayesian models. In {\itshape Proceedings of the }{\itshape Proceedings of
  the ninth international workshop on artificial intelligence and statistics},
  Vol. ~1, 2003.

\bibitem[\protect\citeauthoryear{Tuia
  {\itshape{et~al.}}}{2011}]{tuia2011multioutput}
Tuia, D., Verrelst, J., Alonso, L., P{\'e}rez-Cruz, F. and Camps-Valls, G.,
  Multioutput support vector regression for remote sensing biophysical
  parameter estimation. {\itshape IEEE Geoscience and Remote Sensing Letters},
  2011, \textbf{8}, 804--808.

\bibitem[\protect\citeauthoryear{Vapnik}{2000}]{vapnik2000nature}
Vapnik, V., {\itshape The nature of statistical learning theory}, 2000,
  Springer.

\bibitem[\protect\citeauthoryear{Vazquez and Walter}{2003}]{vazquez2003multi}
Vazquez, E. and Walter, E., Multi-output support vector regression. In
  {\itshape Proceedings of the }{\itshape 13th IFAC Symposium on System
  Identification}, pp. 1820--1825, 2003.

\bibitem[\protect\citeauthoryear{Yang {\itshape{et~al.}}}{2010}]{yang2010high}
Yang, D., Liang, G., Jenkins, D.D., Peterson, G.D. and Li, H., High performance
  relevance vector machine on GPUs. In {\itshape Proceedings of the }{\itshape
  Symposium on Application Accelerators in High-Performance Computing}, 2010.

\end{thebibliography}
